\documentclass{article}

\usepackage{arxiv}

\usepackage[utf8]{inputenc} 
\usepackage[T1]{fontenc}    
\usepackage{hyperref}       
\usepackage{url}            
\usepackage{booktabs}       
\usepackage{amsfonts}       
\usepackage{nicefrac}       
\usepackage{microtype}      
\usepackage{lipsum}
\usepackage{graphicx}
\graphicspath{ {./images/} }
\usepackage{xcolor}
\usepackage{multirow}

\usepackage{textcomp}
\usepackage{gensymb}
\usepackage{amsmath}
\DeclareMathOperator*{\argmax}{arg\,max}

\usepackage{graphicx}
\DeclareGraphicsExtensions{.pdf,.png}

\title{Potential Impacts of Smart Homes on Human Behavior: A Reinforcement Learning Approach}

\author{
 Shashi Suman$^1$, Ali Etemad$^1$, Francois Rivest$^2$ \\
  $^1$Dept. ECE and Ingenuity Labs Research Institute, Queen's University\\
  $^2$Dept. of Mathematics and Computer Science, Royal Military College of Canada\\
  \texttt{\{shashi.suman, ali.etemad\}@queensu.ca, francois.rivest@\{mail.mcgill.ca, rmc.ca\}}
  }

\begin{document}
\maketitle
\begin{abstract}
Smart homes are becoming increasingly popular as a result of advances in machine learning and cloud computing. Devices such as smart thermostats and speakers are now capable of learning from user feedback and adaptively adjust their settings to human preferences. Nonetheless, these devices might in turn impact human behaviour. To investigate the potential impacts of smart homes on human behavior we simulate a series of Hierarchical-Reinforcement Learning-based human models capable of performing various activities namely setting temperature and humidity for thermal comfort inside a Q-Learning-based smart home model. We then investigate the possibility of the human models' behaviors being altered as a result of the smart home and the human model adapting to one another. For our human model, the activities are based on Hierarchical-Reinforcement Learning. This allows the human to learn how long it must continue a given activity and decide when to leave to pursue a different one. We then integrate our human model in the environment along with the smart home model and perform rigorous experiments considering various scenarios involving a single human model and two human models with the smart home. Our experiments show that with the smart home, the human model can exhibit unexpected behaviors like frequent changing of activities and an increase in the time required to modify the thermal preferences. With two human models, we interestingly observe that certain combinations of models result in normal behaviours, while other combinations exhibit the same unexpected behaviours as those observed from the single human experiment.

\end{abstract}

\section{Introduction}

Smart Home Systems (SHS) have drastically evolved in recent years due to advances in AI, enhanced connectivity, affordability, and the Internet of Things (IoT). For instance, smart thermostats \cite{SmartThermostat} and smart device schedulers \cite{LightMusic} are increasing in popularity everyday. These devices and the ecosystems which they create, aim to enhance human quality of life by saving time and costs, and by increasing comfort. On the other hand, the increased use of AI and automation in our daily lives has begun to impact human performance \cite{AutomationHumanPerformance}. For instance, it was shown in \cite{ProductivityHappiness} that with increased efficiency in human performance, which can be the result of automation and the use of AI, happiness also increases. 

\begin{figure}[t]
    \centering
    \includegraphics[width=0.5\columnwidth]{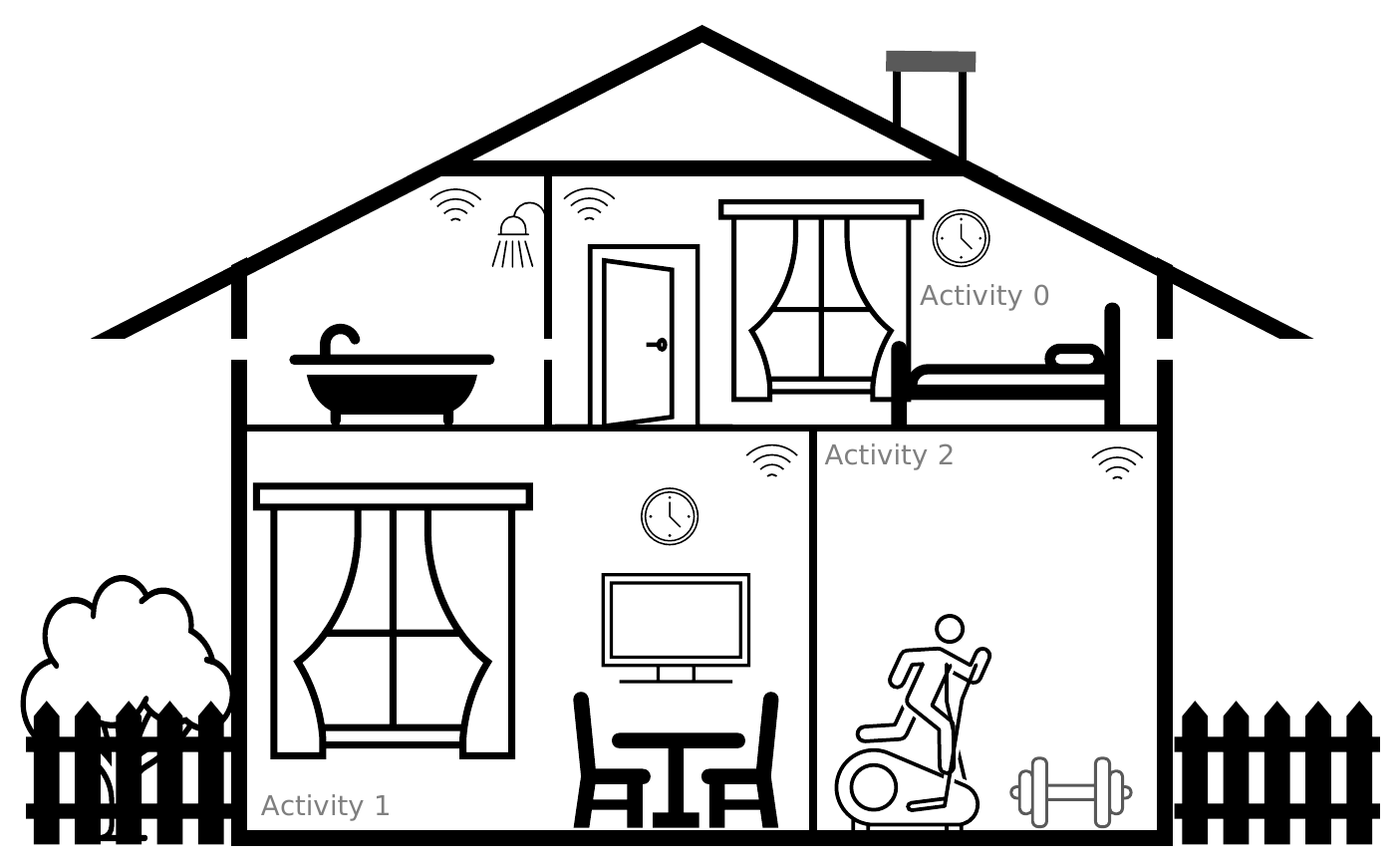}
    \caption{A home equipped with an SHS (Signal symbol) in each room and a single human model that pursues 3 activities where \textit{Activity 0} is rest indicating light activity, \textit{Activity 1} is watching TV indicating medium activity, and \textit{Activity 2} is physical workout indicating heavy activity.}
   \label{fig:Home_SHS_Single}
\end{figure}

Reinforcement learning (RL) has been widely used to enable agents to learn from feedback provided either by other agents or the environment \cite{BartoSutton}. As a result, it has been extensively used in intelligent/automated agents \cite{IntelligentAgents}. 

While RL-based agents generally learn to optimize their performance for maximizing received rewards, the maximized received reward may only be optimal from the perspective of the agent without achieving the goal of the environment. For example, it was shown in \cite{AtariExploitation} that a smart agent can exploit the environment to gain significant rewards by taking advantage of an edge-case in the rules of the environment without completing the game.

In recent years, SHS have also begun to utilize RL for learning from human interactions and feedback in order to provide customized and personalized user experience \cite{SmartLight} \cite{SmartThermostat}. While intelligent agents such as SHS generally provide value and increase quality of life, \textit{we question whether it is possible for an SHS powered by RL to learn to exploit the intricacies of human behaviours in order to maximize its own reward regardless of its implications on the humans}. 

In this paper, we tackle this problem by investigating if it is possible for an adaptive RL-based smart home to exploit assumptions in the environment and consequently \textit{change the behaviors of the human agents} while trying to maximize its received reward by controlling thermal comfort parameters. To do so, we run a set of experiments on an RL-based simulated smart home and human agents in three steps. In the first step, we train a Hierarchical Reinforcement Learning based human model to pursue a set of activities while also setting temperature and humidity (TH) parameters to gain maximum activity-based rewards and thermal comfort rewards. In the second step, we train the smart home model with our trained human model. Here, the SHS learns to anticipate the thermal preference of the human model for each activity from the received feedback. In the final step, we integrate the trained human model into the smart home, and run/evaluate them together. Here, the SHS receives the temperature, humidity, and activity as observations, and sets the temperature and humidity in order to provide maximum comfort to the human model for that activity, based on its learned policy. For the smart home to learn to adapt to human preferences based on activities, we use Q-Learning \cite{HumanFeedback}. For the human model, we utilize \textit{Hierarchical Reinforcement Learning}, which is capable of efficiently modeling the ability of the human agent to switch between activities \cite{BOTVINICK}. Figure \ref{fig:Home_SHS_Single} shows an overview of our study. Our simulations show that indeed unintended consequences can be caused in certain scenarios, as we observe behavioral changes exhibited by the human agent such as frequently switching between the activities along with increased time-steps to set the TH settings. We also perform multi-human experiments where two human agents are simultaneously integrated into a single home. Interestingly, we observe that in this scenario, human models are able to change their task order to minimize the differences between their TH preferences as to reduce the time spent in changing TH and maximize their comfort.

In summary, our contributions in this paper are as follows.
(\textbf{1}) We model a smart home with RL to learn to optimize ambient parameters, namely temperature and humidity for maximizing human agents' comfort. We then model a simulated human agent capable of pursuing and switching between various activities. The agent can also control the temperature and humidity levels in order to manage its own comfort. 
(\textbf{2}) Through our simulations, we find that in an environment that follows well-established physics laws of heating and humidity, the co-existence of the RL-based SHS and the human model can lead to unexpected changes in the optimal policy that would otherwise not be present without the SHS, resulting in a change in the behaviour of the human model.
(\textbf{3}) We then perform additional experiments to consider a scenario where two human models simultaneously pursue their activities and try to optimize their comfort in parallel in a home environment. We discover that unintended behaviors are present with models that have less overlapping comfort ranges and different reward structures.

\section{Related Work} \label{RelatedWork}
In this section, we review some of the key papers in three main areas related to this paper. First, we review papers that use AI in smart environments. Next, we explore prior works in the area of human-RL interaction. These two surveys are necessary as our study integrates AI in the form of RL in a simulated smart home environment and aims to analyse the interaction between RL-based SHS and simulated human agents. Lastly, we review past works that have used RL to model human behaviour as we aim to simulate humans in the smart home using RL. It is therefore critical to base our human models on accepted approaches that have used RL for modeling human agents.

\textbf{AI in Smart Environments.} Recent advances in machine learning have enabled smart devices to learn user behaviors and preferences for enhanced user experience and comfort. For instance, in \cite{SmartThermostat} smart thermostats were capable of learning user thermal preferences in real-time using Bayesian networks. In \cite{BehaviorSmartHome}, neural networks were used to predict user behaviors in smart homes to adjust the ambient thermal conditions. 

RL algorithms have also begun to make their way into SHS. In \cite{SmartLight}, value iteration, a reward-based algorithm, was used to learn user preferences about lighting, conditional on the number of occupants, time of day, and surrounding illumination. Multi-agent deep Q-Networks have been used in \cite{EnergyDQN} to learn the energy cost of smart home appliances. Deep RL was employed in \cite{EnergyDDPG} for energy minimization along with a neural network that predicted user thermal comfort. This was further extended in \cite{EnergyAirDQN} where air quality for a particular temperature was controlled using the Double Deep Q-Network (Double DQN) algorithm while also minimizing the energy consumption. Similarly in \cite{AirQualityQLearning} Q-Learning was used to control the indoor ventilation to maintain air quality by learning the rate of CO$_2$ generation from individuals. An autonomous ventilation system was designed in \cite{airqualityDQN} that learned to maintain the indoor air quality using DQN in a simulated environment, managing to reduce the energy consumption as well. Similarly in \cite{airqualityDistributedRL}, an air quality sensing system was developed where robots sensed the air quality via sensors and optimized their navigation path using Partial Observable Markov Decision Process (POMDP) and Double DQN.

\textbf{Human-RL Interaction.}
A number of studies have explored the interaction between humans and RL systems. For instance, \cite{HumanIntervention} discusses the notion of placing humans in the loop to provide feedback while an agent plays Atari games. With this approach, the learning stability of the agent increased as the number of iterations required for convergence was reduced. Similarly in \cite{HumanRLConversation}, bots learned to converse through feedback from humans using the Reinforce algorithm, resulting in an improved learning curve.

Based on studies such as the above, it can be observed that including humans in RL loops generally results in better AI learning. Nonetheless, some studies have interestingly discovered potential negative or unintended consequences of human-RL interactions. For instance, it was shown in \cite{HumanIntervention} that in complex environments where the human is unsure about the nature of feedback it has to provide to the agents, AI learning frequently fails. Moreover, in some other studies, it has been shown that RL agents can exploit certain rules in the environment to maximize their reward regardless of its implications on the overall system. For example in \cite{AtariExploitation} the agent exploited the environment to obtain indefinite rewards, while in \cite{openaiExploitation}, the agent exploited the physics engine of the environment to win the game.

\textbf{RL for Modeling Humans.} Modeling human behavior as an artificial agent is a challenging task due to the existence of a large number of factors that govern our behaviors. As a result, its underlying learning mechanisms and associated reward systems are still open research problems. However, since the discovery of links between dopamine and the reward prediction error signal in RL \cite{rewardPredictionTheoryDopamine}, RL models have become key in studying multiple facets of human behaviours including learning, motivation, addiction, decision making, and numerous diseases \cite{montague2004computational}. RL models of learning, behaviours, and dopamine are now well established and used in a large number of publications \cite{KEIFLIN2015247}.

Authors in \cite{HumansHRL} showed hierarchical RL to be more accurate in contextualizing and generalizing tasks, and measuring variations in performance between the users when compared with flat RL. In \cite{humanDecisionHRL}, hierarchical RL showed high accuracy in sentiment classification, noise reduction, and prioritizing certain sentiments in a given text document. Authors in  \cite{SubtasksHRL} \cite{SubtasksHRL_2} \cite{SubtasksHRL_3} showed that humans in general divide a given task into sub-tasks and solve them in a hierarchical order that would return the highest reward at the end. Similarly, authors in \cite{RLinBrain} talked about the presence of hierarchical structures in the brain while making decisions, suggesting that we possess a cognitive control of sub-tasks based on their positions in the hierarchy order. These studies, among others, back the plausibility of using HRL-based models for simulating human decision making and behaviour in interaction with smart agents or environments.

\section{Methods} \label{Method}
\textbf{Problem Setup.} We aim to investigate whether for human agent $\mathcal{H}$ capable of interleaving a set of activities through policy $\mathcal{\pi_H}$, there exists an RL-based SHS, $\mathcal{M}$, that can change the human behavior in an attempt to maximize comfort. Specifically, for a policy $\mathcal{\pi_H}$ which $\mathcal{H}$ learns in the absence of $\mathcal{M}$, we aim to obtain $\mathcal{\pi_H}' \neq \mathcal{\pi_H}$ when $\mathcal{M}$ is integrated into the environment. To do so, we assume $\mathcal{M}$ controls TH combined with an existing HRL model for $\mathcal{H}$. We design $\mathcal{H}$ capable of carrying out the activities rest, leisure, and physical workout, which are three common categories of activities in normal human life, through policy $\mathcal{\pi_H}$. $\mathcal{M}$ will learn to anticipate the TH preferences of $\mathcal{H}$, such that $\mathcal{H}$ would feel more comfortable while completing the activities through a different policy $\mathcal{\pi_H}'$. In the next two sections, we describe $\mathcal{M}$ and four variations of $\mathcal{H}$ to evaluate whether our hypothesis on the possibility of obtaining $\mathcal{\pi_H}' \neq \mathcal{\pi_H}$ is correct.

\textbf{Smart Home System.} We model the SHS using a Markov Decision Process (MDP) defined by the tuple $<~\mathcal{S}, \mathcal{A}, P, R, \gamma>$, where $\mathcal{S}$ is the set of discretized states, $\mathcal{A}$ is the set of actions, $P (s'|s,a)$ is the probability to go to state $s'$ when taking action $a$ in state $s$, $R(s,a)$ is the expected reward given when taking action $a$ in state $s$, and $\gamma$ is the discounted factor devaluing rewards received later in time. The objective is to learn a policy $\pi(s)$ which returns the action to take in state $s$ that maximises the expected sum of future rewards given as
$\sum_{t=1}^{\infty}\gamma^t r_t$,
where $r_t$ is the reward received after taking the $t^{th}$ action. The expected sum of future rewards received for taking action $a$ in state $s$ by following policy $\pi$ is given by the Bellman equation: 
\begin{equation}
    Q^{\pi}(s, a) = \sum_{s'\in S} P(s'|s, a)[R(s, a)+\gamma Q^{\pi}(s', \pi(s')].
\end{equation}
The optimal policy $\pi^*(s) = \argmax_{a\in\mathcal{A}}\{Q(s, a)\}$ is obtained using Q-Learning \cite{QLearning}. Given a transition from state $s$ to state $s'$ using action $a$ and reward $r$, the update rule is given by:
\begin{equation}
    Q(s, a) \leftarrow (1-\alpha)Q(s, a) + \alpha \left[r + \gamma\max_{a'\in\mathcal{A}}\{Q(s', a')\}\right],
\end{equation}
where $\alpha$ is the learning rate. While learning, actions are selected using a decaying $\epsilon$\nobreakdash-Greedy policy to keep a balance between exploration and exploitation. The $\epsilon$\nobreakdash-Greedy policy is given by 
\begin{equation}
\pi(s) = \begin{cases}
    \argmax_{a\in\mathcal{A}}\{Q(s, a)\} \text{ with probability } 1-\epsilon\\
    a\in\mathcal{A} \text{ with probability } \epsilon/|\mathcal{A}|
\end{cases}.
\end{equation}

Based on the $\epsilon$ value, the agent decides between a random versus a greedy action, where the value of $\epsilon$ decays with each iteration. 

The smart home receives state $s$ that consists of temperature, humidity, and the current activity that our human model is pursuing as a discrete state. Given that many recent wearable devices that track \textit{human activity} can now be integrated into smart-home applications, we include the human activity in the observation that the smart home receives as a state to ensure the smart home can incorporate the activity when setting the temperature thus learning preference of the model for a specific activity. Based on this observation, the smart home may take an action to increase, maintain, or decrease the temperature or humidity respectively. The smart home receives a negative reward (-1) every time the human changes the TH, thus, encouraging the smart home to set the TH to the human preference.

To simulate realistic changes in temperature and humidity of the environment, we use thermal calculations based on Newton's law of cooling \cite{Newton} assuming the dew-point to be constant because of low deviation in temperatures to keep Newton's law from breaking. If we consider our Human temperature preference to be $T_h$ and the surrounding temperature at time $t$ as $T_t$, then the temperature at the next time-step can be expressed as:
\begin{equation}
    T_{t+1} = T_h + (T_{t} - T_h)e^{-k},
\label{eg:Temp}
\end{equation}
where $t$ is the number of time-steps and $k=0.8$ is the decay constant that can be used to decide how fast or slow the desired temperature is achieved. For simplicity, we ignore the heat generated by the human body's metabolism which varies due to sweat evaporation, radiation, convection, and conduction. Hence, we keep the mean radiant temperature at 22 degrees to reduce the complexity.

To simulate the changes in humidity within the environment when the user makes changes in the temperature, we use the August-Roche-Magnus approximation \cite{TempHumidityRelation}. As per definition, relative humidity in terms of dry bulb temperature and dew-point temperature can be approximately expressed as:
\begin{equation}
    R_h = 100 \times \frac{e^{\frac{17.625\times T_d}{243.04+T_d}}}{e^{\frac{17.625\times T}{243.04+T}}},
\end{equation}
where $T$ is the room temperature (Eq.~\eqref{eg:Temp}) and $T_d$ is the dew-point temperature. Similarly, when the human agent or the smart home sets the humidity, changes in temperature can be defined approximately as:
\begin{equation}
    T = 243.04 \times \frac{{\frac{17.625\times T_d}{243.04+T_d}} - \ln{\frac{R_h}{100}} }{17.625 + \ln{\frac{R_h}{100}} - \frac{17.625\times T_d}{243.04+T_d}},
\end{equation}
where $T$ is the approximate temperature in \degree C given current humidity $R_h$ and dew-point temperature $T_d$.
Since most homes are thermally insulated, very little variation is observed in temperature and humidity, making the dew-point fairly stable. In this paper, the simulated temperature varies between 15$\degree$C and 30$\degree$C. The dew-point does not vary much within this range, thus we consider a fixed dew-point of $T_d = 4\degree C$ for our simulations.


\textbf{Human Thermal Comfort.}
To represent human thermal comfort, we use the Fanger \cite{standard199255} model that provides an empirical solution. One of the important parameters described by the model is called the Predicted Mean Vote (PMV) giving an understanding of the human sensation towards thermal comfort. This is the mean response for a large group of people for various thermal conditions. The PMV index spans from -3 to 3 with 3 being the hottest feeling, -3 being the coldest feeling, and 0 being the most comfortable feeling. PMV depends on six factors, which are the dry-bulb temperature, relative humidity, clothing index, wind speed, and metabolism index (specific to each activity). Based on \cite{standard199255}, the clothing index for rest, leisure, and physical workout is 0.5, 0.67, and 0.36 respectively. Similarly, we have set wind speed in the surroundings to be 0. The optimal PMV range for the best comfort lies between -0.5 to 0.5. For our human model, we have used the PMV equations which can be found in \cite{standard199255} to express the intrinsic reward to the human agent.

\textbf{Human Basic Model.} 
We need a human model $\mathcal{H}$ that can: (\textit{i}) represent approximately how humans adapt based on reward; (\textit{ii}) carry out or switch between tasks; and (\textit{iii}) set TH to gain comfort and reward. Accordingly, we use reinforcement learning for the human model as such approaches have frequently been used for modeling human behaviors and adaptation based on rewards \cite{HumanReinforcementDopamine} \cite{BOTVINICK} \cite{gebhardt} \cite{BOTVINICK2014} \cite{RLdopamine}. In particular, an HRL model capable of capturing human behaviors in selecting, completing, and switching among different tasks has been proposed in \cite{gebhardt} \cite{BOTVINICK} \cite{BOTVINICK2014}.

\begin{figure}[t]
    \centering
    \includegraphics[width=0.5\columnwidth]{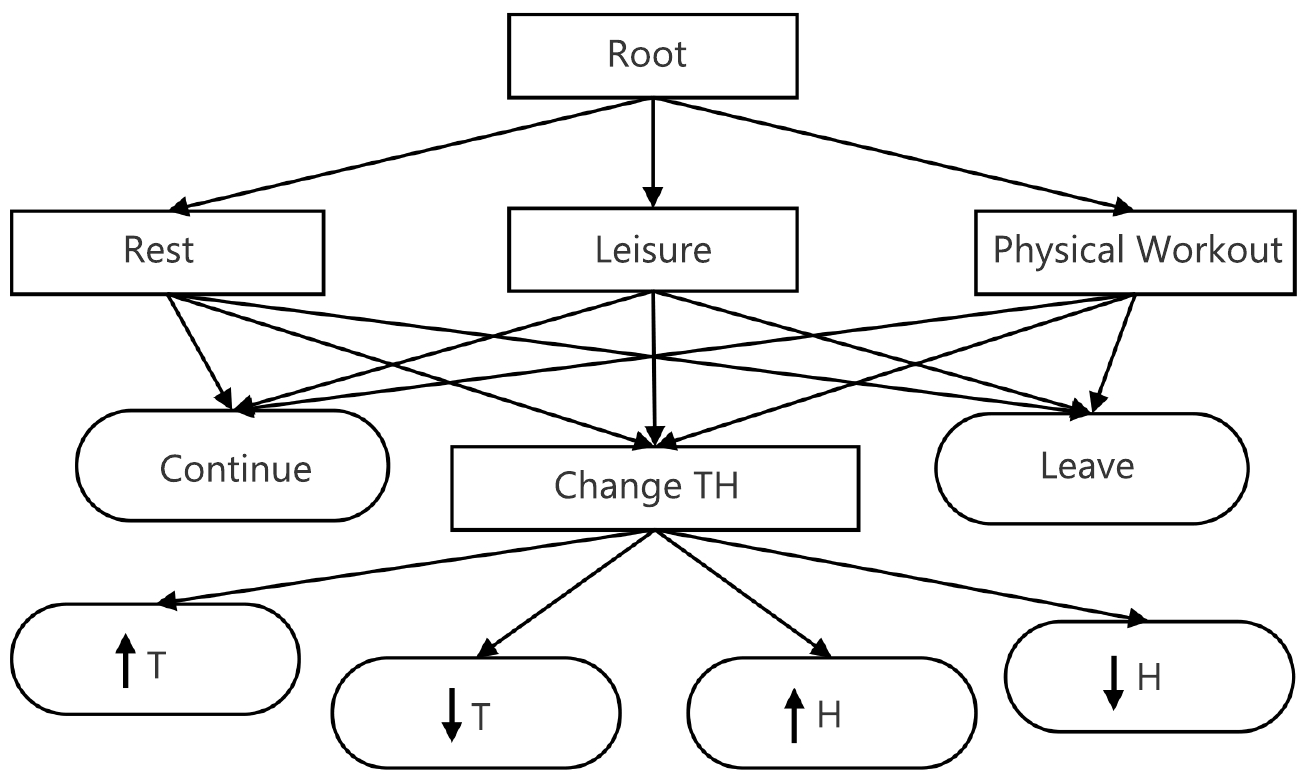}
    \caption{Hierarchical model representation for our simulated human. Rectangles are composite SMDP subroutines and ovals are primitive actions. Here T stands for \textit{Temperature} and H stands for \textit{Humidity}.}
    \label{fig:Human HRL Model}
\end{figure}

We design $\mathcal{H}$ such that within a limited time window $N$, it learns to select or leave the available or ongoing activities to maximize reward. We use a discounted Hierarchically-MAXQ (HMAXQ) algorithm from \cite{Andre}, which was shown to simulate well-known human behavior of task-interleaving \cite{gebhardt}. In this formulation, the goal can be broken down into a hierarchy of sub-problems where each sub-problem is a separate Semi-Markov Decision Process (SMDP) that the human model has to learn. This allows us to combine primitive actions to create more composite ones. We select three possible types of activity classes based on activity intensity for the human agent: `rest' for low intensity, `leisure' for medium intensity, and `physical workout' for high intensity. Each activity class has a different cost function, state-space, and PMV preferences. The entire hierarchy has four composite actions and six primitive (non-SMDP) actions as shown in Figure \ref{fig:Human HRL Model}. An SMDP is similar to an MDP, with the exception that some actions can take multiple time-steps to be completed, and hence the agent receives delayed rewards. This is due to the fact that some actions are represented as macros consisting of sequences of actions.

We divide the hierarchy for $\mathcal{H}$ into three distinct levels. The first level is the \textit{root} node SMDP. The root subroutine chooses an activity that returns a maximum cumulative reward. The second level consists of tasks that the human agent must perform. Each task is an SMDP subroutine that models the progress within an activity based on the time spent in it so far. This helps us to generalize the interleaving ability of humans within activities. In each activity, the human can either \textit{continue} or \textit{leave} the activity, or decide to change either the temperature or humidity as the third level in the hierarchy.

\begin{figure}[t]
    \centering
    \includegraphics[width=0.5\columnwidth]{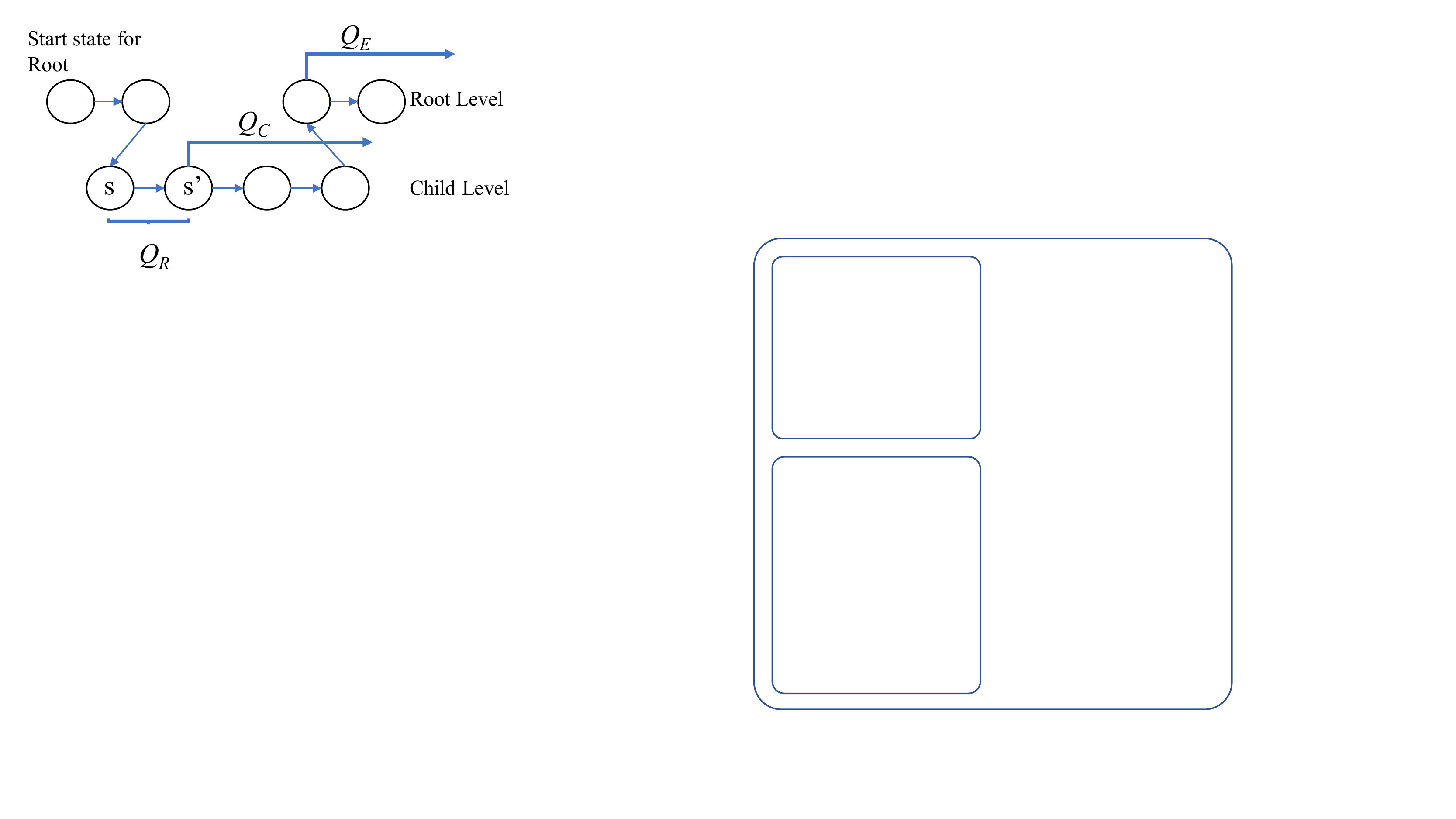}
    \caption{Hierarchical Representation the decomposed $Q$ function, where $Q_r$ is the immediate reward for primitive action, $Q_c$ is the reward to complete the subroutine, and $Q_e$ is the expected reward from external subroutines. $s$ is the current state and $s'$ is the next state.}
    \label{fig:HRL Representation}
\end{figure}

To model our HMAXQ-based human agent, we use a three-part value function decomposition as described in \cite{Andre, gebhardt} and given by:
\begin{equation}
Q_{r}(a, s) = 
\begin{cases}
    \sum_{s'} P(s'|s, a)R(a,s') 
    \qquad \textit{if $a$ is primitive} \\
    Q_{r}(child(i), s) +
    Q_{c}(child(i), s, \pi_{child(i)}(s)) \\
    \qquad\qquad\qquad\qquad\qquad \ \ \ \textit{if $a=i$ is composite}
\end{cases}
\label{5}
\end{equation}
\begin{equation}
\begin{split}
    Q_{c}(i, s, a) = \sum_{SS}P'(s', t|s, a)\gamma^{t}[Q_{r}(\pi_{i}(s'), s') \\ + Q_{c}(i, s', \pi_{i}(s'))]
\end{split}
\label{6}
\end{equation}
\begin{equation}
Q_{e}(i, s, a) = 
    \sum_{EX}P'(s', t|s, a)\gamma^{t}Q^{\pi}(p(i), s', \pi_{p(i)}(s')) \\
\label{7}
\end{equation}
where $Q_{r}$ is the expected discounted reward for taking action $a$, $Q_{c}$ is the expected discounted reward for completing the subroutine $i$ after the agent takes action $a$ in state $s$, $Q_{e}$ refers to all the rewards external to the current subroutine (i.e., after exiting the current subroutine), and $p(i)$ is the parent subroutine of $i$. $SS$ are the states that exist within the same subroutine of $s$. $EX$ are the exit states for the current subroutine where the human can leave the task. The value of $\gamma$ in Eq.~\ref{6} can be used to control the behavior of our human agent. A smaller value of $\gamma$ models the attractiveness of attaining nearby rewards thus increases switching to an activity that has a high reward. A larger value avoids switching between activities and increases the reward horizon of the agent to complete the activity. Our human model can exit from a task at any given state, hence, every state is considered as an exit state. $i$ indicates the node value of each task in the hierarchy. $0$ is assigned to the root node at level $1$, and $1, 2, 3,...$ are assigned to the composite tasks for the consecutive levels. Figure~\ref{fig:HRL Representation} shows the decomposition of the value function describing the trajectory of $Q$ value for the agent in terms of $Q_r$, $Q_c$, and $Q_e$. 

The full Bellman equation for the the hierarchy is given by:
\begin{equation}
Q^{\pi}(i, s, a) =
\begin{cases}
    Q_{r}(a, s) + Q_{c}(0, s, a) \qquad \textit{if $i=0$ is root} \\
    Q_{r}(a, s) + Q_{c}(i, s, a) +  Q_{e}(i, s, a)  \textit{ otherwise}
\end{cases}
\label{HRLEQ}
\end{equation}
where $\pi$ is the policy. The Bellman equation for the root does not contain $Q_{e}$ because there is no external subroutine at the root level that provides external rewards to the root node.

\textbf{Optimality.} Hierarchical Reinforcement learning is categorized into two notions for an optimal solution. \textit{i}) \textit{Recursive Optimal} and \textit{ii}) \textit{Hierarchical Optimal}. In Recursive Optimal or MAXQ \cite{Dietterich}, the expected return for performing a subtask $a$ is $Q^{\pi}(i, s, a) = Q_{r}(a, s) + Q_{c}(i, s, a)$. The expected external reward $Q_{e}(i, s, a)$ after the completion of subtask $i\neq0$ is not a part of the value function which makes MAXQ a Recursive Optimal solution.  
In contrast, in Hierarchical Optimal \cite{Andre}, when solving the entire hierarchy, the policy is hierarchically optimal given the rewards from the external subroutines (Eq.~\ref{7}). Each task may or may not be locally optimal. The Bellman equation for Hierarchical Optimal is given in Eq.~\ref{HRLEQ}.

In this paper, we have taken $\mathcal{H}$ as a Hierarchical Optimal agent to imitate the switching ability of humans where they have context about the external rewards from other available tasks. The learning rules can be found in \cite{Andre}.

\begin{figure}[t]
    \centering
    \includegraphics[width=0.75\columnwidth]{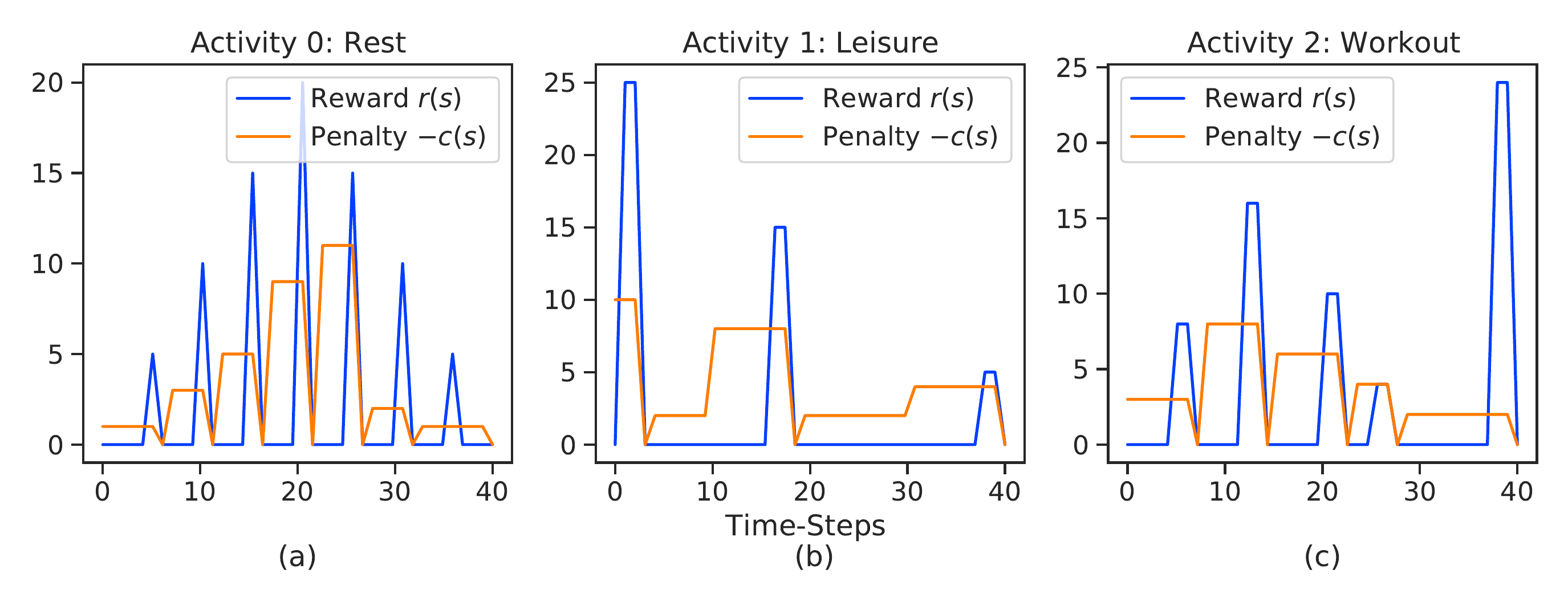}
    \caption{Penalty \textit{-c(s)} and Rewards \textit{r(s)} for leaving or continuing the activity `rest', `leisure', and `physical workout' with respect to the total time spent in the activity.}
    \label{RewardPenalty}
\end{figure}

\textbf{Reward Structure.}
We design the reward structure for all three activity types based on the generalized assumption a given task can be divided into sub-tasks, which in turn can help humans switch between them based on given set of priorities \cite{SubtasksHRL,SubtasksHRL_2}. In \cite{gebhardt} this is modeled by rewarding the time-steps required to complete milestone within an activity, and by penalizing leaving of the activity prior to achieving the milestone. The reward represents the motivation that the human agent senses for a given activity in a given state \cite{wickensAppliedAttention_1}. Similarly, the penalty models the time required for humans to recall the forgotten information after an interruption  \cite{wickensAppliedAttention_2}.

For our simulations, we have three activities, namely, `rest', `leisure', and `physical workout' as shown in Figure~\ref{RewardPenalty}. For the `rest' activity (Activity 0), as supported by \cite{moreSleepMoreProblems}, we model the reward $r(s)$ such that each period of rest increases the reward up to a maximum point, while providing smaller rewards for the subsequent periods (Figure~\ref{RewardPenalty}(a)). For the `leisure' activity (Activity 1), similar to \cite{GotBored,SubtasksHRL_3}, we assume that the reward decreases over time. In the `physical workout' activity (Activity 2), spending time on workout increases the performance up to an optimal point, after which it starts decreasing as modeled by \cite{stressPerformance}. Occasionally, individuals can extend their workout and go beyond the fatigue threshold to achieve a considerable milestone and obtain a high reward as shown in \cite{motivationBeyondFatigue,motivationBeyondFatigue_2} and modeled in our reward function (Figure~\ref{RewardPenalty}(c)).

While pursuing any of the activities, it is important for the human model to show reluctance in switching between tasks and make an effort to continue the current activity.
To elaborate, in \cite{wickenstastSwitch}, it was demonstrated that the task-switching ability of humans is based on cognitive factors like task attractiveness, priority, engagement, interest, and others, which are difficult to model since each person might express these feelings differently. We model this by penalizing the `leave' action (Figure~\ref{RewardPenalty}), as well as the root to enforce reluctance in frequent task-switching. The root is penalized with a reward which is given as: 
\begin{equation}
    R(0, s, a) = -c(task(s)),
\end{equation}
where $task(s)$ returns the state of the child SMDP corresponding to root state $s$, and $c(s)$ is the penalty for leaving the task. Figure \ref{RewardPenalty} shows the penalty (in red) for leaving the task or activity.

Finally, we integrate a penalty to represent discomfort when the TH is outside the comfort zone, encouraging the model to make adjustments when needed. Our final reward structure for the human model $\mathcal{H}$ is given by:
\begin{equation}
    R_{1}(s, a) = 
    \begin{cases}
        r(s) - d & \text{if $a$ is \textit{continue}} \\
        -c(s) & \text{if $a$ is \textit{leave}},
    \end{cases}
    \label{H_A_Reward_Equationn}
\end{equation}

Since we want to show that differences between the expected human model used when training the SHS and the actual human model could lead to issues, we define a second reward function for $\mathcal{H}$ as:
\begin{equation}
R_{2}(s, a) = 
    \begin{cases}
        r(s) - d & \text{if $a$ is \textit{continue}} \\
        -c(s) + d & \text{if $a$ is \textit{leave}} \\
    \end{cases}
    \label{H_C_Reward_Equation}
\end{equation}
where $d$ is defined as:
\begin{equation}
d_{0.5} = 
    \begin{cases}
        |pmv(s)| & \text{if $|pmv(s)|$ $>$ $0.5$} \\
         0 & \text{if $|pmv(s)|$ $<$ $0.5$} \\
    \end{cases}
    \label{Thermal_Score}
\end{equation}

Here $pmv(s)$ is the intrinsic comfort reward given for the current state $s$, while $r(s)$ and $c(s)$ are the reward and penalty functions for pursuing and leaving the activity respectively as shown in Figure~\ref{RewardPenalty}. In both equations $pmv(s)$ returns the thermal reward of the state as given in Eq.~\ref{Thermal_Score}. In Eq.~\ref{H_C_Reward_Equation}, it represents the sense of relief when humans leave an uncomfortable environment for a more comfortable one. In the next section, we will use Eq.~\ref{H_A_Reward_Equationn} and Eq.~\ref{H_C_Reward_Equation} to represent the reward function of various human agents.

\section{Experiments and Results} \label{Experiments and Results}
In this section, we describe our experiments and present the results. First, we utilize only one human agent in a smart home, while in the second sub-section, we integrate two human agents simultaneously into a single smart home to perform the studies for a multi-human setting.

\subsection{SHS with a single Human model in the environment}
We perform the following experiments to evaluate different assumptions regarding the human model $\mathcal{H}$ while interacting with the SHS. We train each human model for 350 episodes, which are empirically enough for convergence. Next, we train the human models with the SHS for an additional 150 episodes. We repeat our experiments \textit{50 times} to evaluate how the SHS performs with different combinations of TH settings when initializing the environment. Following we describe the different human models and experiments along with the obtained results.

First, we explore whether the SHS can learn the preferences of a given general human model. Second, we test whether such SHS could adapt to a human agent with a slightly different set of TH preferences. Finally, we test what would happen if the SHS interacts with a human agent that has a different reward structure than the general human model.

\subsubsection{Experiment 1: Baseline} 
In this experiment, we explore a baseline scenario where an SHS is trained for an average human model using the existing thermal comfort sensational model \cite{standard199255}. Here, we aim to show that (1) our human model can indeed learn to complete each available activity, (2) the SHS can learn to anticipate the human TH preferences, and (3) the human performance does not exhibit any unexpected behaviors when integrated into the environment with the SHS.

Let Model $\mathcal{H}_A$ be a generic human model used to train the SHS with the optimal comfort PMV indices (\textit{i.e.} the SHS initial settings) between $-0.5$ and $0.5$ as per \cite{standard199255}. The metabolic rates used by the PMV function for the 3 activities are $[1.0, 1.3, 1.8]$ respectively as defined in \cite{standard199255}. The reward function for $\mathcal{H}_A$ is given by Eq. \ref{H_A_Reward_Equationn}.

\begin{figure}[t]
    \centering
    \includegraphics[width=0.65\columnwidth]{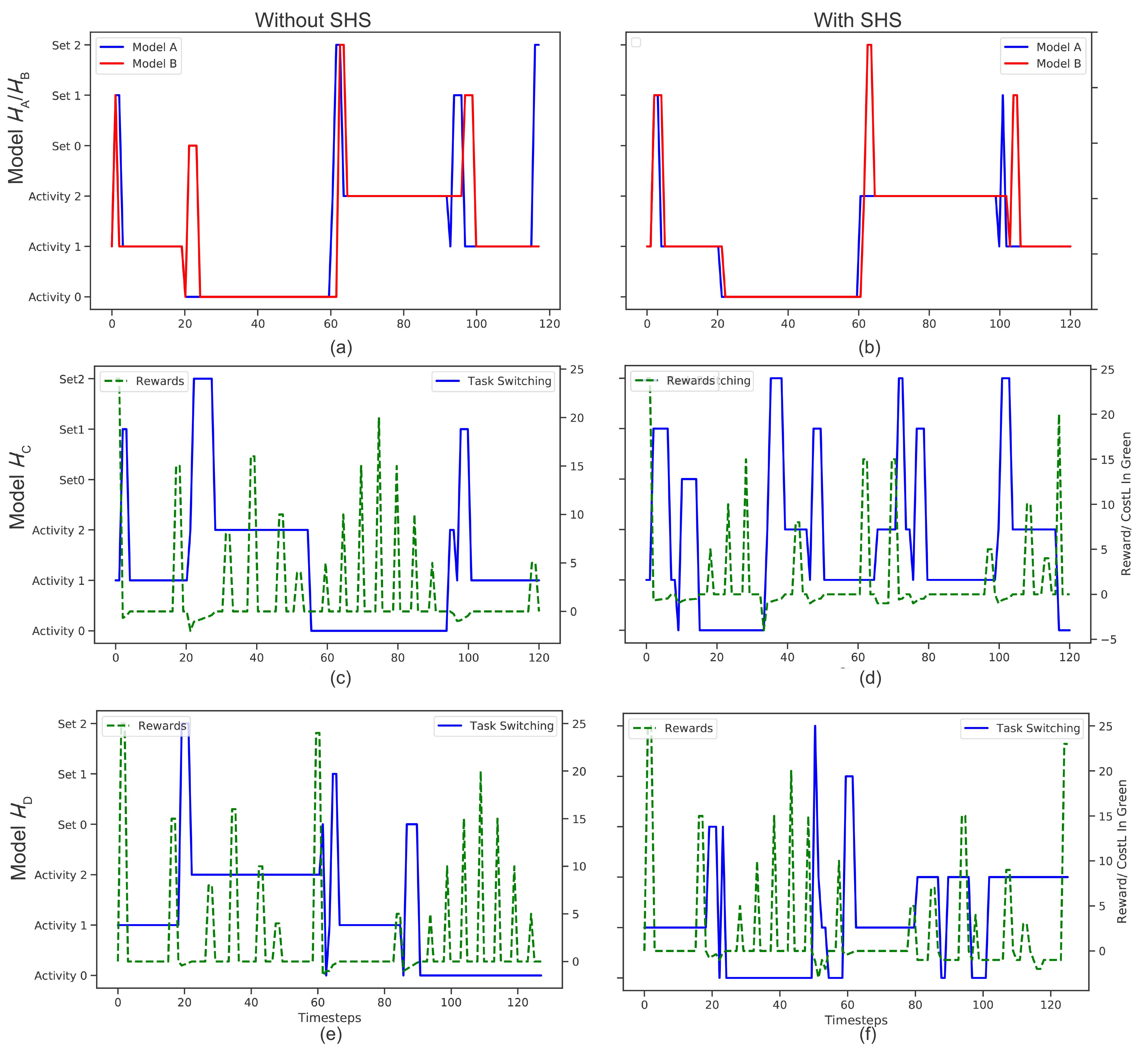}
    \caption{Sample plots of activities over time for each model ($\mathcal{H}_A$, $\mathcal{H}_B$, $\mathcal{H}_C$, $\mathcal{H}_D$) with and without the SHS. In (a), (c), (e), each model learns to complete the tasks without interruption. In (b) and (f), the SHS anticipates human preferences, speeding up the time for human models $\mathcal{H}_A$, $\mathcal{H}_B$, and $\mathcal{H}_D$ respectively to complete the activities. In (d), Model $\mathcal{H}_C$ (different internal reward structure) behaves erratically in the presence of SHS. `\textbf{Set} \textit{i}' denotes the action of setting TH for the corresponding activity \textit{i}.
    }
   \label{Exp_1_2_3}
\end{figure}

As seen in Figures \ref{Exp_1_2_3}(a) and (b), this experiment shows that Model $\mathcal{H_A}$ is capable of completing each activity along with setting the optimal TH settings. As expected, with the SHS, the number of time-steps required for $\mathcal{H_A}$ to correct the TH is reduced as shown in Figure \ref{Exp_1_2_3_timestep}(a), which indicates that the SHS learns to quickly adjust the TH to the human preference for the current task, and the human to expect the SHS in doing so.

\subsubsection{Experiment 2: Control} \label{Control  Experiment}
In this experiment, our goal is to validate that the SHS can learn to adapt to a human model with a thermal preference different to that of our average human model (Model $\mathcal{H}_A$). Here, we define Model $\mathcal{H}_B$ as an agent with different activity-specific TH preferences compared to Model $\mathcal{H}_A$, but with the same reward functions defined by Eq. \ref{H_A_Reward_Equationn}. The metabolic values within the PMV for the three activities of Model $\mathcal{H}_B$ are set to $[1.10, 1.35, 1.75]$ based on \cite{standard199255}, which also changes the TH preference as the activity-specific metabolism rates are associated with different TH values for optimal comfort.

In Figure \ref{Exp_1_2_3}(a), we can see that both human models $\mathcal{H}_A$ and $\mathcal{H}_B$, once trained, converge to the expected behaviors by completing all 3 activities, and spending a few time-steps tuning the TH at the beginning of each activity. Then, Figure \ref{Exp_1_2_3}(b) shows that once the SHS is integrated into the environment, the amount of time spent by the human model in adjusting the TH is reduced, without negatively affecting the completion of activities by the human agent. As expected, with the SHS, the human model takes fewer time-steps to correct the TH. Figure \ref{Exp_1_2_3_timestep}(a) and (b) show that for the human agent with the SHS (in green), the time-steps are reduced as compared to the human model without the SHS (in red) for both Models $\mathcal{H}_A$ and $\mathcal{H}_B$. This concludes that the SHS can learn to adapt to Model $\mathcal{H_B}$ with a different thermal preference than that of $\mathcal{H_A}$. This works well for the TH preferences of an average model $\mathcal{H}_A$ as well as for $\mathcal{H}_B$ with different preferences, showing that the SHS model can adapt to both generic and specific human model preferences. With Experiments 1 and 2, we can conclude that the human models in the RL-based smart home can successfully complete the activities without abruptly leaving any of them while requiring fewer time-steps to set TH as shown in Figure \ref{Exp_1_2_3_timestep}.

\subsubsection{Experiment 3: Demonstration} In this experiment, we look into the prospect of the SHS learning the preference of a human that has a different intrinsic thermal reward structure than the ones in Models $\mathcal{H}_A$ and $\mathcal{H}_B$. Here we introduce two human models namely Model $\mathcal{H_C}$ and $\mathcal{H_D}$, where both models differ from $\mathcal{H}_A$ and $\mathcal{H}_B$ by having an extra reward term for leaving uncomfortable thermal conditions \cite{BOTVINICK}. We define the reward structures of Models $\mathcal{H}_C$ and $\mathcal{H}_D$ by Eq. \ref{H_C_Reward_Equation}.

The metabolism indices of Model $\mathcal{H_C}$ are set to $[1.80, 1.35, 1.15]$ for activity 0, activity 1, and activity 2 respectively which is moderately different from the baseline (Model $\mathcal{H_A}$). Similarly, for Model $\mathcal{H_D}$, we set it to $[1.15, 1.25, 1.85]$ for each activity respectively which is similar to the baseline model. This changes the TH preference for Model $\mathcal{H}_D$ during different activities. We are here considering the case where human models with different reward function might have similar or different thermal preferences than the baseline model. This helps to learn about the factors that may possibly impact human model's behavior in the presence of SHS. The comfortable PMV range for both models is between -0.25 to 0.25 ($d_{0.25}$ Eq.~\ref{Thermal_Score}) to simulate a more constrictive range of PMV for our human agents.

\begin{figure*}[t]
    \centering
    \includegraphics[width=1\textwidth]{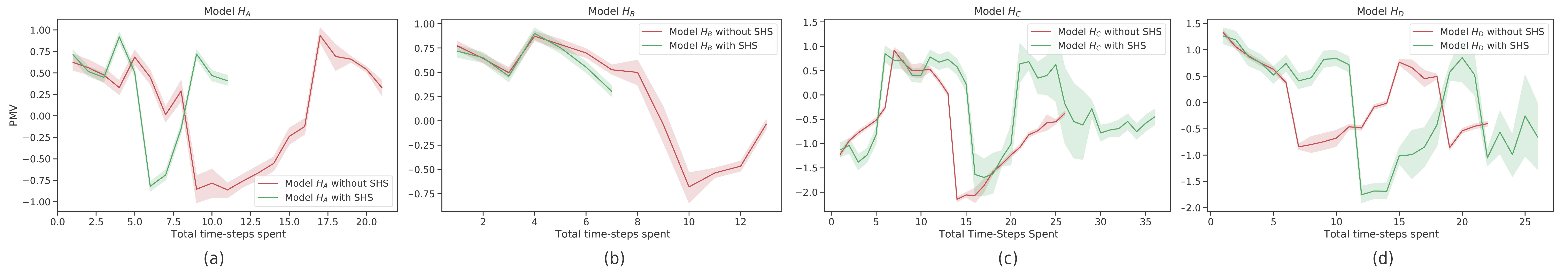}
    \caption{PMV trajectory of human models $\mathcal{H}_A$, $\mathcal{H}_B$, $\mathcal{H}_C$, and $\mathcal{H}_D$ while setting TH for comfort, with and without the SHS. Each colored plot denotes PMV variations of all activities versus the required time-steps to reach the optimal PMV range while pursuing the activities. {\color{red}{Red}}: without the SHS, {\color{green}Green}: with the SHS.}
    \label{Exp_1_2_3_timestep}
\end{figure*}

\begin{figure}
    \centering
    \includegraphics[width=0.5\columnwidth]{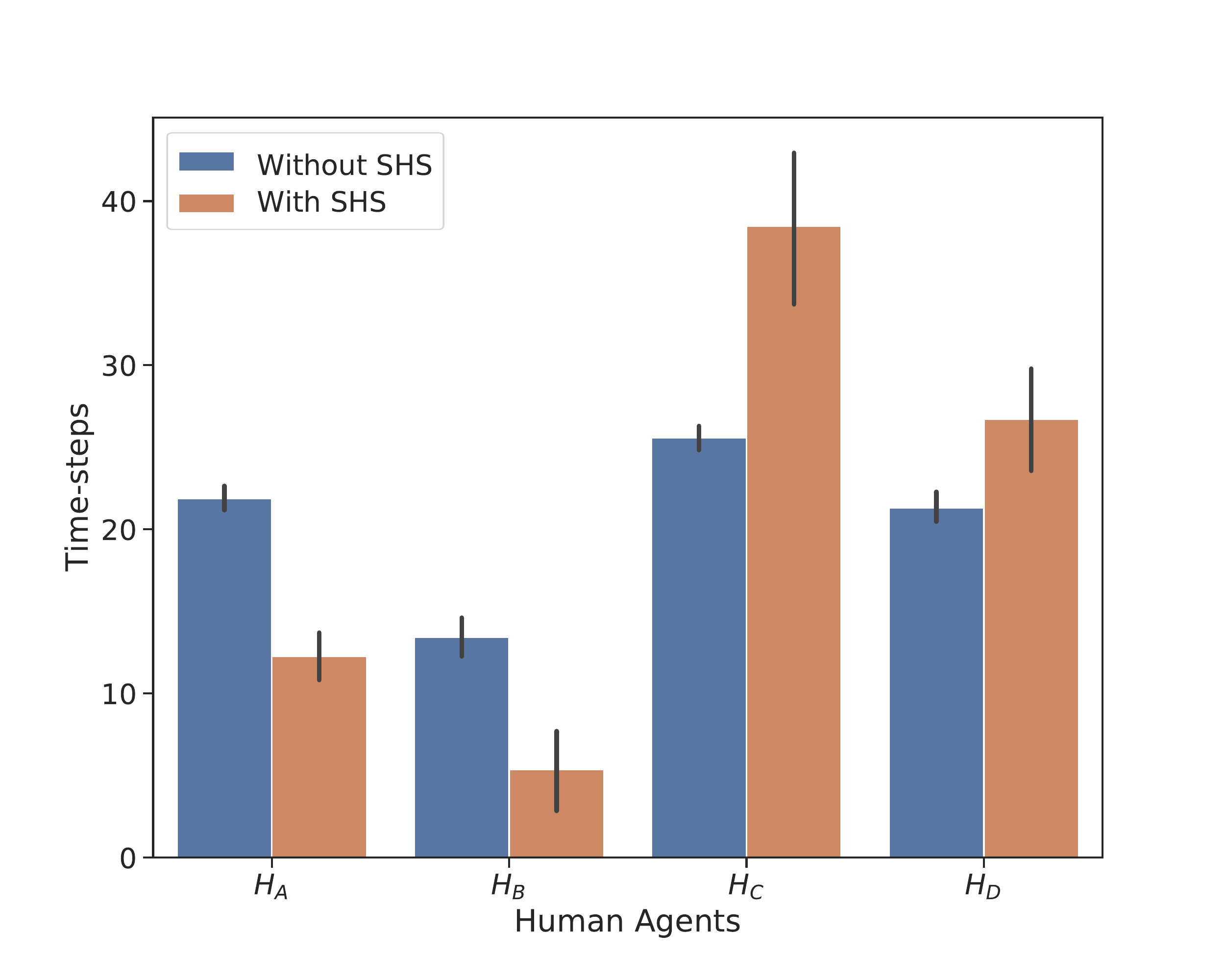}
    \caption{Time-steps of human Models $\mathcal{H}_A$, $\mathcal{H}_B$, $\mathcal{H}_C$, and $\mathcal{H}_D$ while setting TH for comfort, with and without the SHS.}
    \label{Exp_All_BarPlot}
\end{figure}

\begin{figure}[t]
    \centering
    \includegraphics[width=0.75\columnwidth]{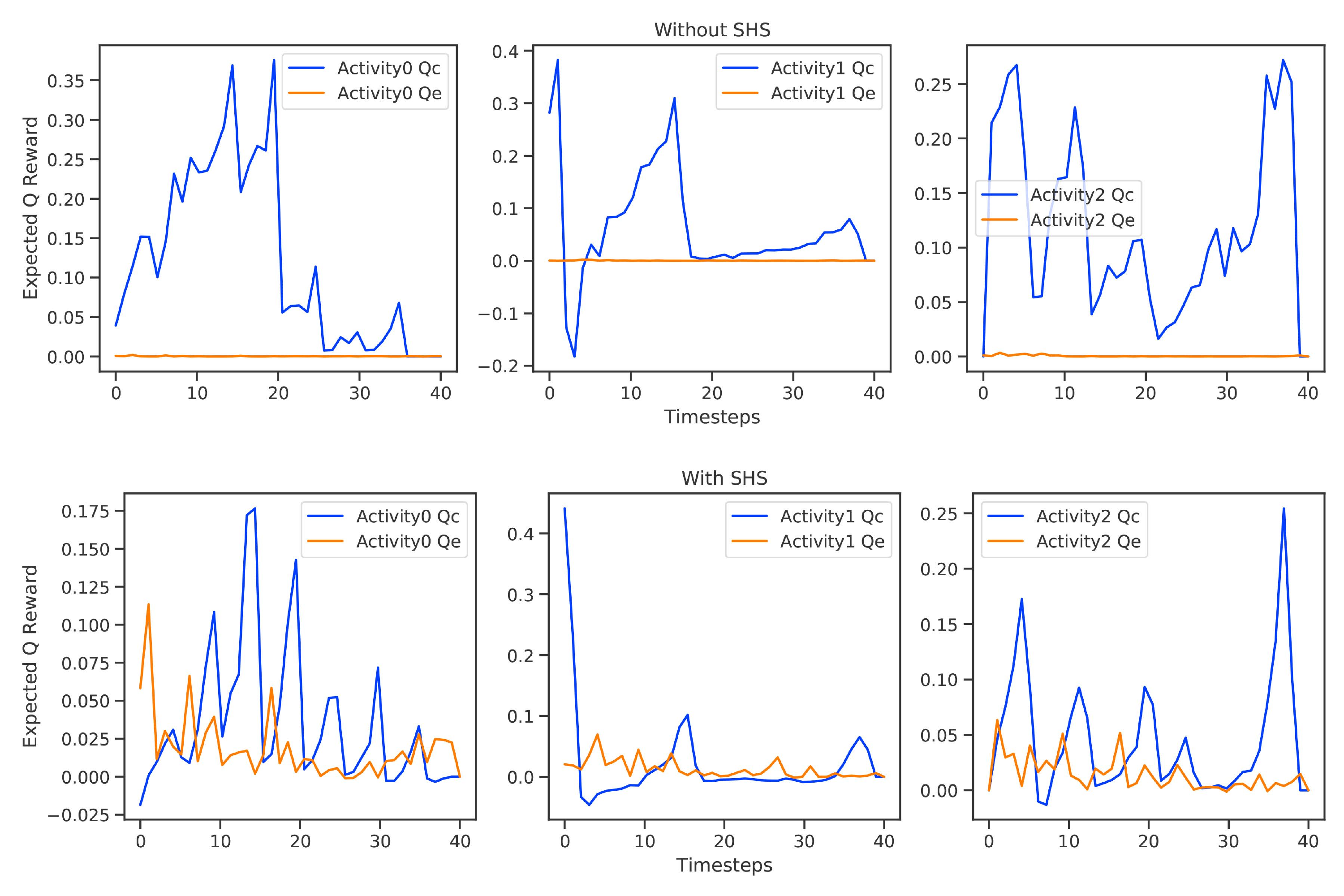}
    \caption{Experiment 3 Model $\mathcal{H}_C$ Expected Q values to complete (\textit{Q\textsubscript{c}}) the current subroutine in blue and expected reward from external subroutines (\textit{Q\textsubscript{e}}). When the expected sum of rewards from external subroutines ($Q_e$ in orange) are higher than the expected reward to complete the subroutine ($Q_c$ in blue), the model exits the current subroutine.}
    \label{Exp_3_QcQe}
\end{figure}

Figure \ref{Exp_1_2_3}(c) (in blue) shows the experiment results where Model $\mathcal{H}_C$ converges to similar behaviors as Models $\mathcal{H}_A$ and $\mathcal{H}_B$ without the smart home. However, with the smart home, Model $\mathcal{H}_C$ deviates from the expected behavior (which is completing the activities without abruptly leaving any of them, as well as setting TH with reduced time-steps). Figure~\ref{Exp_1_2_3_timestep}(c) (in green) shows that with the SHS, Model $\mathcal{H}_C$ takes more time to set the TH than without the SHS (in red). While trying to complete the tasks, we also observe that $\mathcal{H}_C$ occasionally leaves the activities before completing them (Figure \ref{Exp_1_2_3}(d)).

To study this deviation in behavior, we look into the $Q$ \nobreakdash values of $\mathcal{H}_C$ as shown in Figure~\ref{Exp_3_QcQe} and Table \ref{table:H_C_Mean_QcQe}. We observe that the overall reward available in external subroutines ($Q_e$) increases, making $\mathcal{H_C}$ seek rewards from other activities instead of continuing the current activity. Parallel to this, the expected reward to continue within the same subroutine ($Q_c$) decreases in the presence of SHS indicating less enthusiasm to continue and more eagerness to exit to gain rewards from other activities. From Table \ref{table:H_C_Mean_QcQe}, we can observe a decrease in the mean $Q_c$ and an increase in the mean $Q_e$ for each activity with the SHS when compared to without the SHS, indicating the availability of higher rewards from external activities (subroutines). For any state, if the $Q_e$ value is higher than the $Q_c$ value, Model $\mathcal{H}_C$ decides to leave the current activity to maximize the rewards by pursuing external activities. Model $\mathcal{H}_C$ switches to other activities between subtasks since it involves no penalty. Subtasks are small achievements within an overall activity as shown in Figure~\ref{RewardPenalty}. The time-step where our penalty is zero marks the completion of the subtask within an activity. Moreover, activity switching, loss in activity performance, and loss in effort (similar to $Q_c$) to continue with incorrect temperature and humidity were also observed in \cite{thermalEffects}. 

\begin{table}[t]
    \centering
    \caption{Mean $Q_c$ and $Q_e$ values of Model $\mathcal{H}_C$ for each activity with and without the SHS.}
    \begin{tabular}{|c|c|c|c|c|}
   \hline
    & \multicolumn{2}{c|}{Without SHS} & \multicolumn{2}{c|}{With SHS} \\
      & {$Q_c$} & {$Q_e$} & {$Q_c$} & {$Q_e$}\\
   \hline
    Activity 0 & 0.119 & 0.0003 & 0.034 & 0.018\\  
    Activity 1 & 0.063 & 0.0002 & 0.028 & 0.012\\
    Activity 2 & 0.117 & 0.0005 & 0.045 & 0.014\\ 
    \hline
    \end{tabular}
    \label{table:H_C_Mean_QcQe}
\end{table}

\begin{table}[t]
    \centering
    \caption{Mean $Q_c$ and $Q_e$ values of Model $\mathcal{H}_D$ for each activity with and without the SHS.}
    \begin{tabular}{|c|c|c|c|c|}
   \hline
    & \multicolumn{2}{c|}{Without SHS} & \multicolumn{2}{c|}{With SHS} \\
      & {$Q_c$} & {$Q_e$} & {$Q_c$} & {$Q_e$}\\
   \hline
    Activity 0 & 0.1298 & 6.34e-05 & 0.0304 & 0.0184\\  
    Activity 1 & 0.0467 & 7.57e-05 & 0.0234 & 0.0161\\
    Activity 2 & 0.1642 & 5.50e-05 & 0.0186 & 0.0186\\ 
    \hline
    \end{tabular}
    
    \label{table:H_D_Mean_QcQe}
\end{table}

With the occasional switching between activities, the SHS fails to learn the TH preference of Model $\mathcal{H}_C$ during a particular activity which inherently increases the overall time-steps needed to learn the TH preference of the human model. After evaluating Model $\mathcal{H}_C$ (see Table \ref{Exp_123_MTS_MR}, we can see that for 50 trials, the mean reward for Model $\mathcal{H}_C$ has decreased from 264 to 197 showing a decrease in the performance of SHS to learn the preference of Model $\mathcal{H}_C$ with different intrinsic rewards and metabolism indices. Meanwhile, the standard deviation increased from 3.3 to 18.2, similar to our observation in Figure \ref{Exp_1_2_3_timestep}(c), which may suggest additional instability in the overall system.

Next, we compare Model $\mathcal{H}_C$ with Model $\mathcal{H}_D$ that has similar metabolism indices for the activities as the baseline model $\mathcal{H}_A$. Figure \ref{Exp_1_2_3}(e) shows that without the SHS, Model $\mathcal{H}_D$ is able to complete each task and set optimal TH settings. Figure \ref{Exp_1_2_3}(f) shows the progress of Model $\mathcal{H}_D$ successfully completing each activity in the presence of SHS with occasional switching. Figure \ref{Exp_1_2_3_timestep}(d) shows the change in the PMV while completing all activities versus the required time-steps. As can be seen, with the SHS (green plot), the number of time-steps required to set the TH is reduced only by a small margin for Model $\mathcal{H}_D$ indicating that a different reward function impacts the performance of the human model as well as the SHS. With the SHS, for $\mathcal{H_D}$, we observe a decrease in the expected Q-value to continue the activity ($Q_c$) and an increase in the expected Q-value available from external activities ($Q_e$) (see Table \ref{table:H_D_Mean_QcQe}). Evaluations of Model $\mathcal{H}_D$ (see Table \ref{Exp_123_MTS_MR}) shows that with the SHS, the mean reward increase while the time-steps decrease. 

With the above experiments, we arrive at the following conclusions.
\begin{itemize}
    \item Our smart home model can successfully learn to anticipate thermal preference of a given model (i.e., $\mathcal{H_A}$), which does not show an anomaly in the presence of smart home. When integrated with the SHS, $\mathcal{H_A}$ spends fewer time-steps to change the TH without any unintended behaviour like excessive switching between activities.
    \item When a human model with a slightly different thermal preference ($\mathcal{H_B}$) is introduced in the environment, the SHS can learn and adapt according to its preference, thus also managing to reduce the number of time-steps needed to changed the TH.
    \item When a model with a different reward function and different TH preference ($\mathcal{H_C}$) is integrated into the environment, it exhibits an unexpected behavior by frequently switching between activities, resulting in an increase in time-steps.
    \item When a model with a different reward function but similar TH preference ($\mathcal{H_D}$) compared to the baseline model is integrated into the SHS, frequent switching between activities is also observed, and the time-steps to set the TH are relatively high. 
\end{itemize}

\begin{table}[t]
    \centering
    \caption{Mean Time-Steps (MTS) required to change TH, Mean Reward (MR) $\pm$ standard deviation for each model with and without SHS. }
   \begin{tabular}{|c|c|c|c|c|}
   \hline
    & \multicolumn{2}{c|}{Without SHS} & \multicolumn{2}{c|}{With SHS} \\
      & {MTS} & {MR} & {MTS} & {MR}\\
   \hline
    Model $\mathcal{H}_A$ & 21 & 253 $\pm$ 3.2 & 11 & 260 $\pm$  5.6\\  
    Model $\mathcal{H}_B$ & 13 & 255 $\pm$ 4.3 & 08 & 256 $\pm$  9.7\\
    Model $\mathcal{H}_C$ & 26 & 264 $\pm$ 3.3 & 36 & 197 $\pm$18.2\\ 
    Model $\mathcal{H}_D$ & 22 & 255 $\pm$ 5.6 & 26 & 201 $\pm$  12.3\\
    \hline
    \end{tabular}
    
    \label{Exp_123_MTS_MR}
\end{table}

\begin{table}
\centering
\caption{Example of Possible Temperature and Humidity combination in PMV range [-0.25, 0.25] for Model $\mathcal{H}_C$ with metabolism rate of 1.0.}
\begin{tabular}{ |c|c| } 
 \hline
 Temperature $^{\circ}$C & Humidity \% \\
 \hline
    25 & 65\\ 
    25 & 70\\
    26 & 30 \\
    26 & 35 \\
    26 & 40 \\
    26 & 45 \\
    26 & 50 \\
    26 & 55 \\
    26 & 60 \\
    26 & 65 \\
    26 & 70 \\
    27 & 30 \\
 \hline
\end{tabular}
\label{Combination2}
\end{table}

\subsection{SHS with Multiple Human models in the Environment}
In the previous experiments, we explored the impacts of SHS on human performance in an environment with a single human agent. To further extend the above experiments and explore the behavioral performance of multi-human models with and RL-based SHS, we also consider a case of two human models pursuing their activities in the same environment. Our aim with these experiments is to explore whether the SHS can learn the preferences of (a) two human models with similar thermal preferences; (b) two human models with different intrinsic reward and thermal preferences; (c) two human models with same reward function but different from (a); and (d) two human models with different intrinsic rewards but similar thermal preferences.

We first train the the two HRL-based human models such that each can pursue its activities and set TH to obtain optimal comfort within the same environment. We then integrate the SHS with the trained human models and train it until convergence. It is important to note that the order of activity execution is not fixed for these simulations, which means that each model is free to start with an activity that has a maximum $Q$ value.

\begin{figure}
    \centering
    \includegraphics[width=0.75\columnwidth]{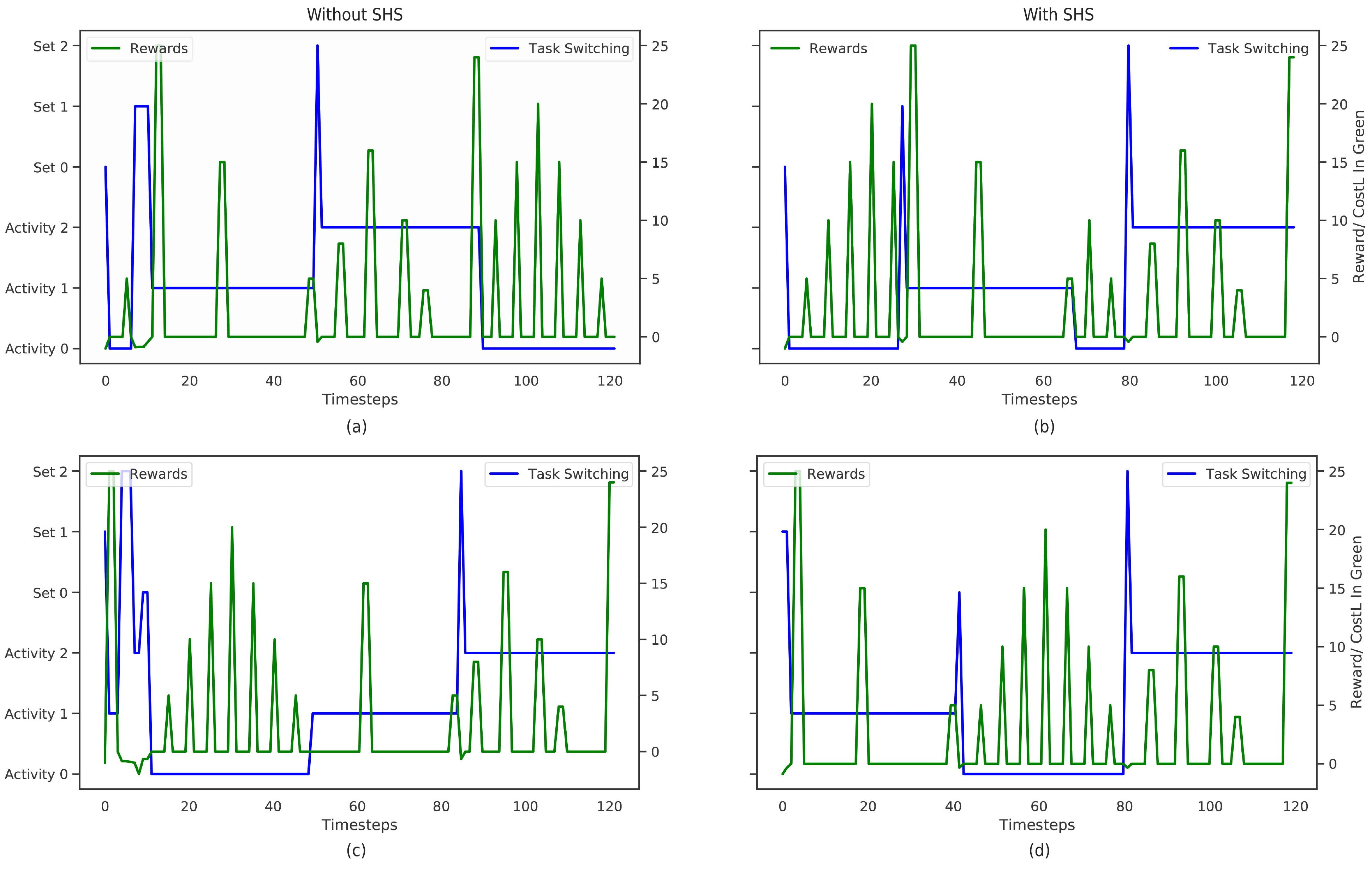}
    \caption{Experiment 4 Top Row: Model $\mathcal{H}_A$, Bottom Row: Model $\mathcal{H}_B$. Activity execution and its corresponding reward (in green). Activity\# is the activity that the human agent is performing (in blue). Set\#: Changing TH for activity \#(in blue).}
    \label{Exp4_All}
\end{figure}

\begin{figure}
    \centering
    \includegraphics[width=0.75\columnwidth]{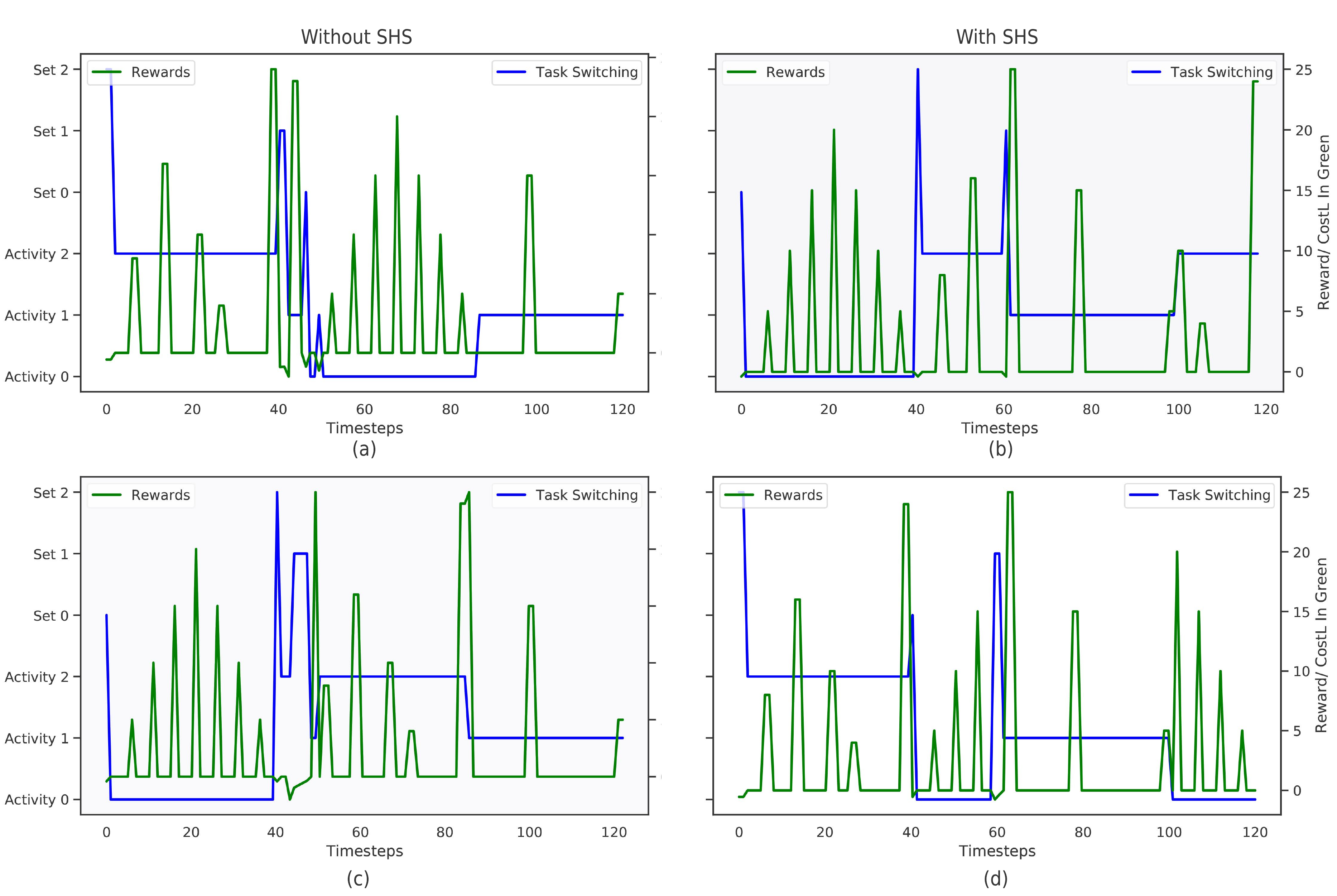}
    \caption{Experiment 5 Top Row: Model $\mathcal{H}_A$, Bottom Row: Model $\mathcal{H}_C$. Activity execution and its corresponding reward (in green). Activity\# is the activity that the human agent is performing (in blue). Set\#: Changing TH for activity \#(in blue).}
    \label{Exp5_All}
\end{figure}

\begin{figure}
    \centering
    \includegraphics[width=0.75\columnwidth]{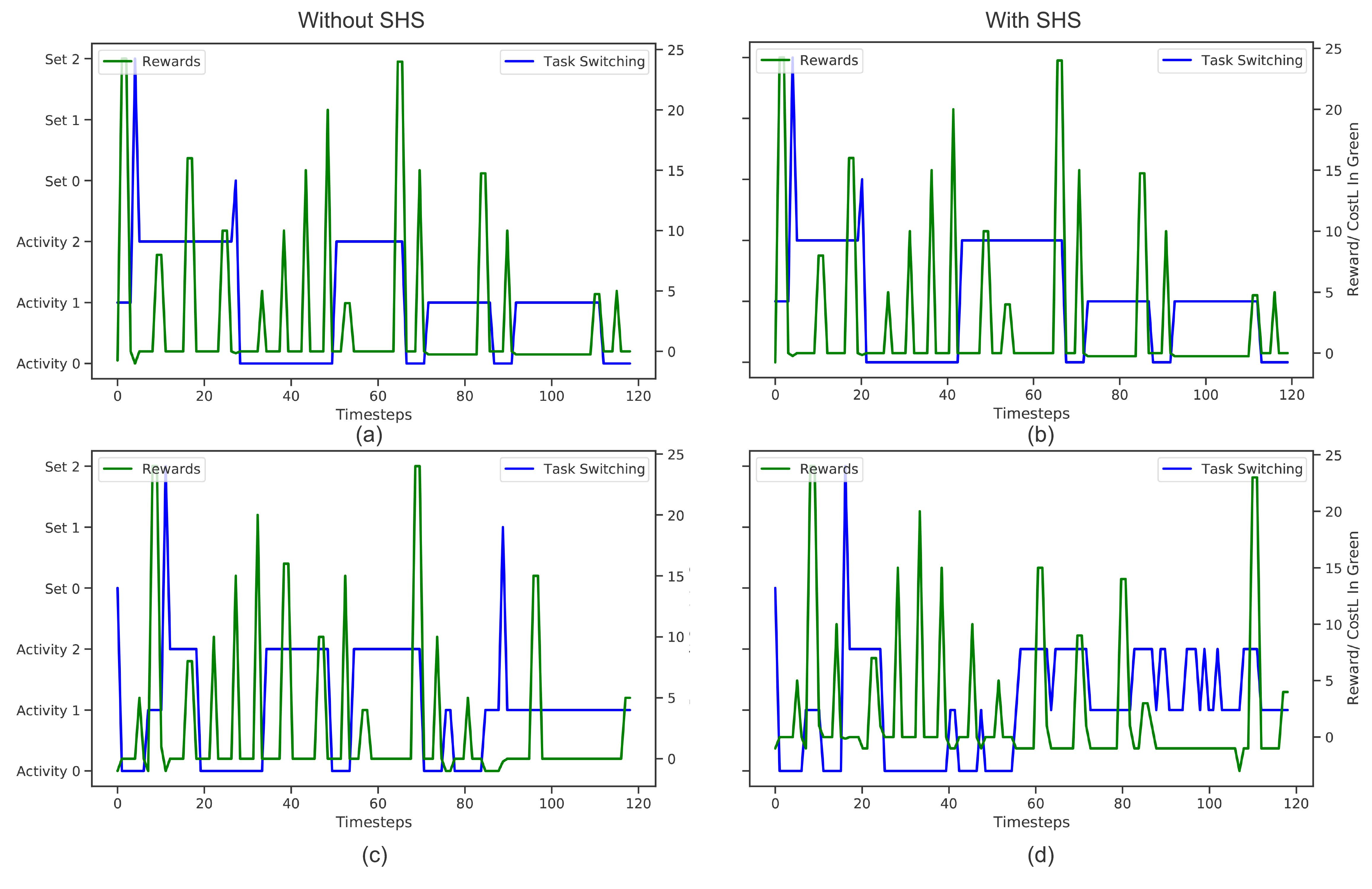}
    \caption{Experiment 5: PMV Range [-0.25, 0.25] Top Row: Model $\mathcal{H}_A$, Bottom Row: Model $\mathcal{H}_C$. Activity execution and its corresponding reward (in green). Activity\# is the activity that the human agent is performing (in blue). Set\#: Changing TH for activity \#(in blue).}
    \label{fig:Experiment_5_Time Steps_0.25}
\end{figure}

\begin{figure}
    \centering
    \includegraphics[width=0.75\columnwidth]{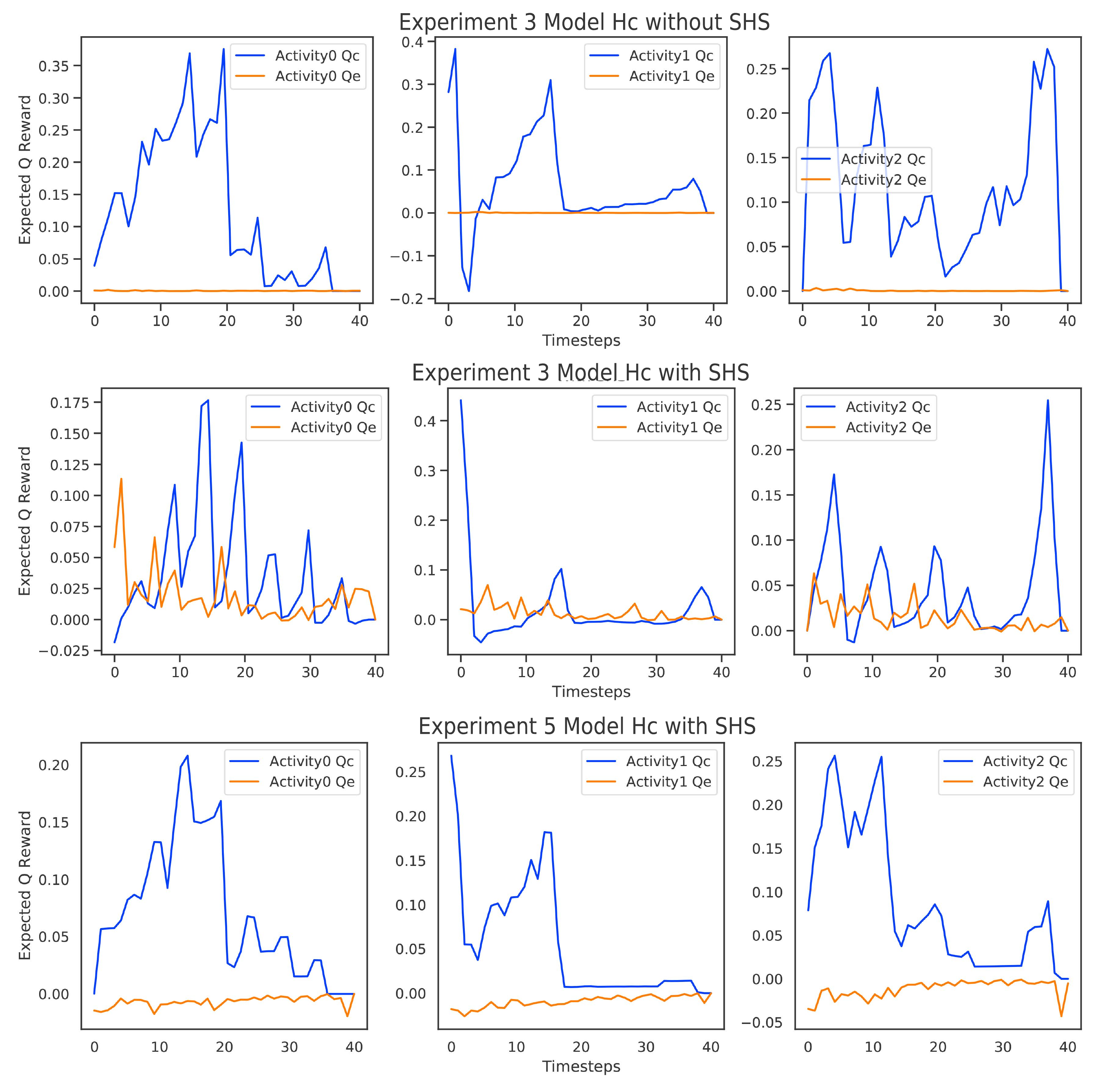}
    \caption{Experiment 5 Model $\mathcal{H}_C$ Expected Q value to complete (in blue) or exit (in orange) the task. In comparison to Experiment 3, with the SHS, the expected Q value to complete the subroutine is higher for $\mathcal{H}_C$ when engaged with $\mathcal{H}_A$.}
    \label{Exp_5_QcQe}
\end{figure}

\begin{figure}
    \centering
    \includegraphics[width=0.5\columnwidth]{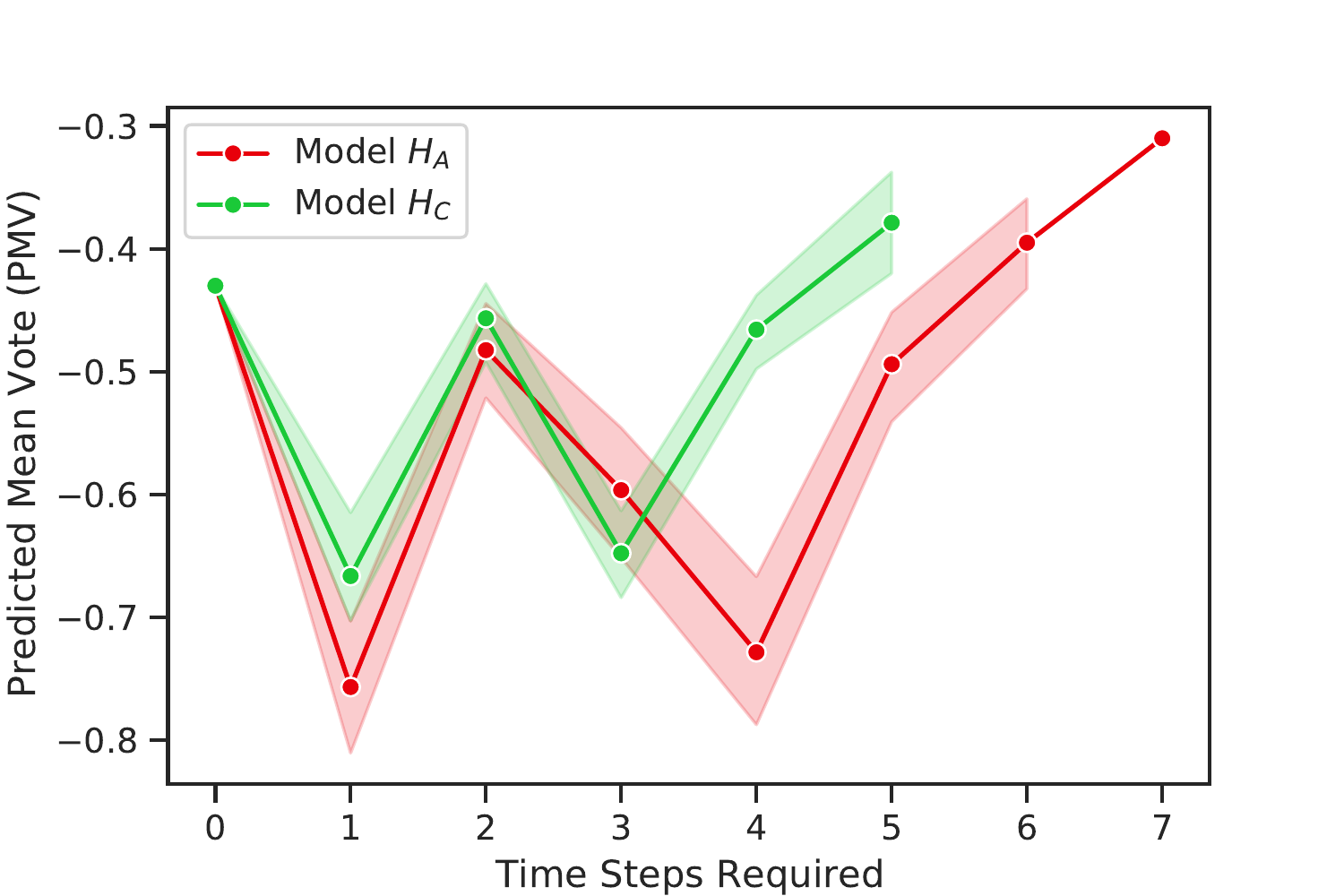}
    \caption{Experiment 5: Similar PMV trajectory of Model $\mathcal{H_A}$ and Model $\mathcal{H_C}$ with the SHS performing activities in parallel and switching between activities with similar thermal comfort to gain maximum rewards. Because of switching between activities having similar thermal comfort, changes in TH made by $\mathcal{H_A}$ benefits $\mathcal{H_C}$ as well.}
    \label{Exp5: Ha_Hc}
\end{figure}

\begin{figure}
    \centering
    \includegraphics[width=0.75\columnwidth]{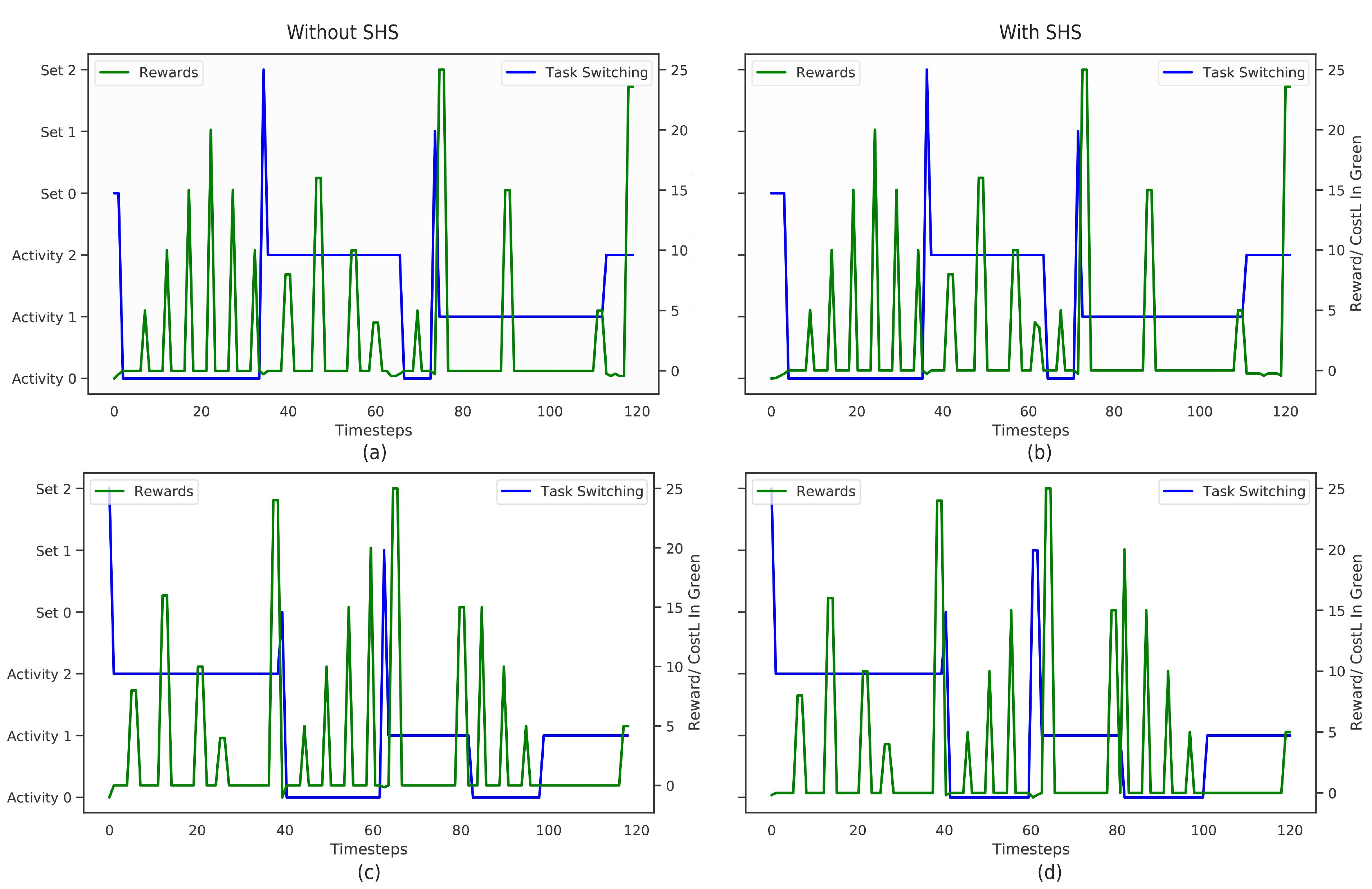}
    \caption{Experiment 6 PMV Range [-0.25, 0.25] Top Row: Model $\mathcal{H}_C$, Bottom Row: Model $\mathcal{H}_C'$. Activity execution and its corresponding reward (in green). Activity\# is the activity that the human agent is performing (in blue). Set\#: Changing TH for activity \#(in blue).}
    \label{Exp6_All}
\end{figure}

\begin{figure}
    \centering
    \includegraphics[width=0.75\columnwidth]{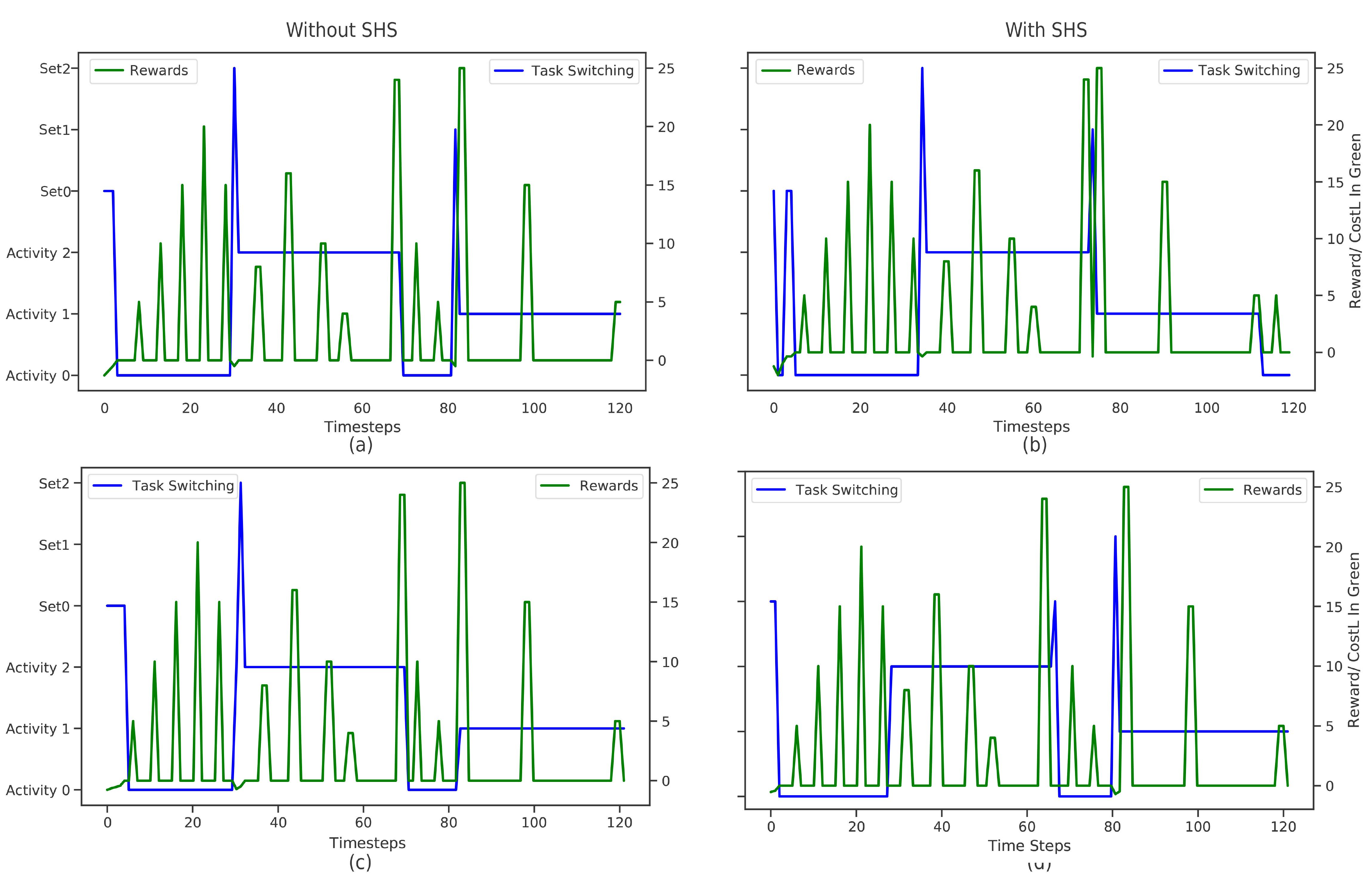}
    \caption{Experiment 7 Top Row: Model $\mathcal{H}_A$, Bottom Row: Model $\mathcal{H}_D$. Activity execution and its corresponding reward (in green). Activity\# is the activity that the human agent is performing (in blue). Set\#: Changing TH for activity \#(in blue).}
    \label{Exp7_All}
\end{figure}

\begin{figure}[t]
    \centering
    \includegraphics[width=0.75\columnwidth]{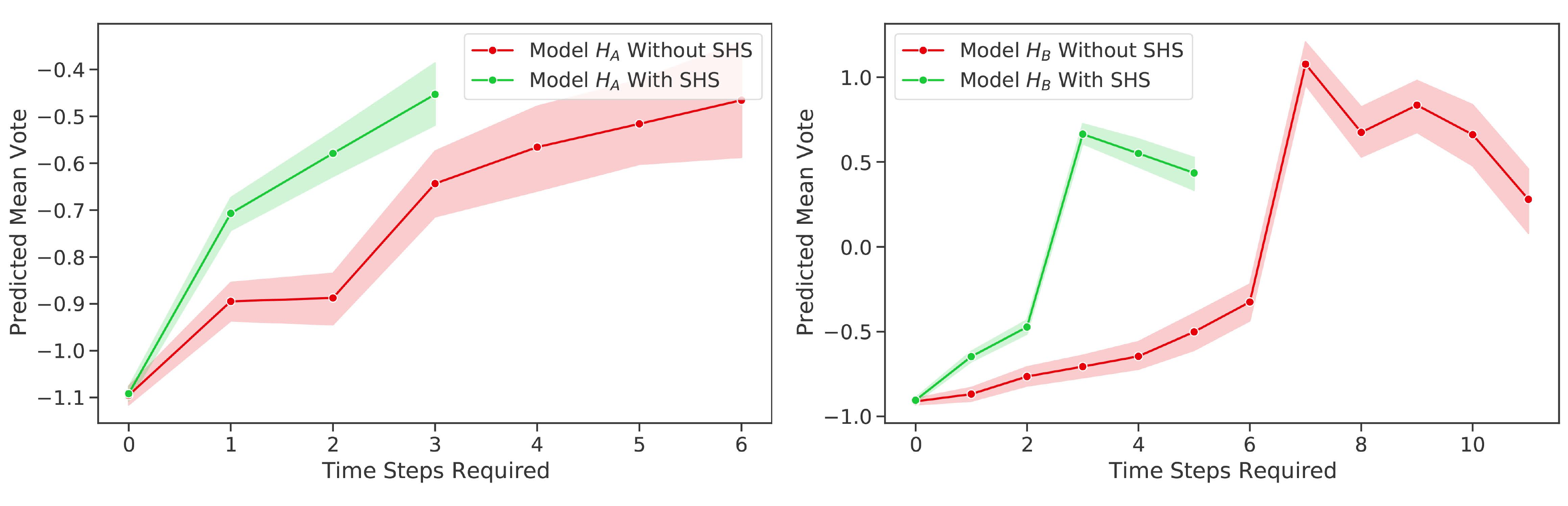}
    \caption{Experiment 4: Model $\mathcal{H_A}$ and Model $\mathcal{H_B}$ PMV trajectory while completing all activities {\color{green}with} and {\color{red}without} the SHS.}
    \label{Exp_4_Timesteps}
\end{figure}

\begin{figure}[t]
    \centering
    \includegraphics[width=0.75\columnwidth]{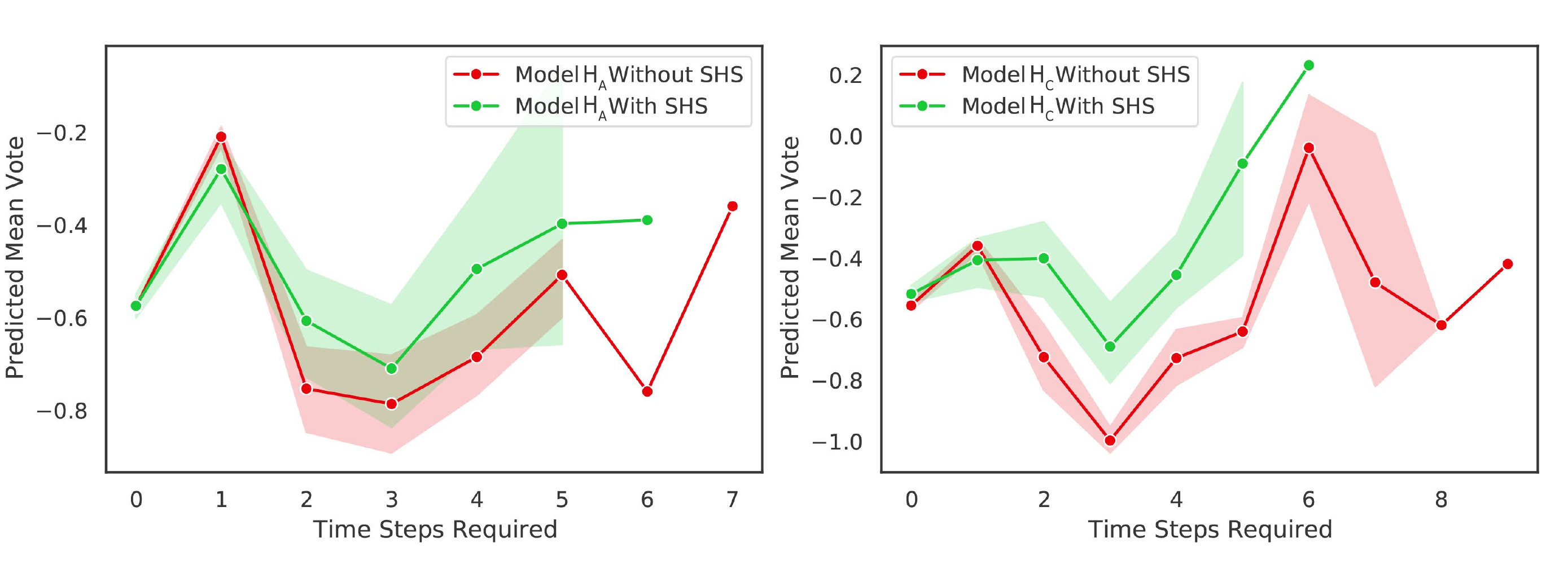}
    \caption{Experiment 5: Model $\mathcal{H_A}$ and Model $\mathcal{H_C}$ PMV trajectory while completing all activities {\color{green}with} and {\color{red}without} the SHS.}
    \label{Exp_5_Timesteps}
\end{figure}

\begin{figure}[t]
    \centering
    \includegraphics[width=0.75\columnwidth]{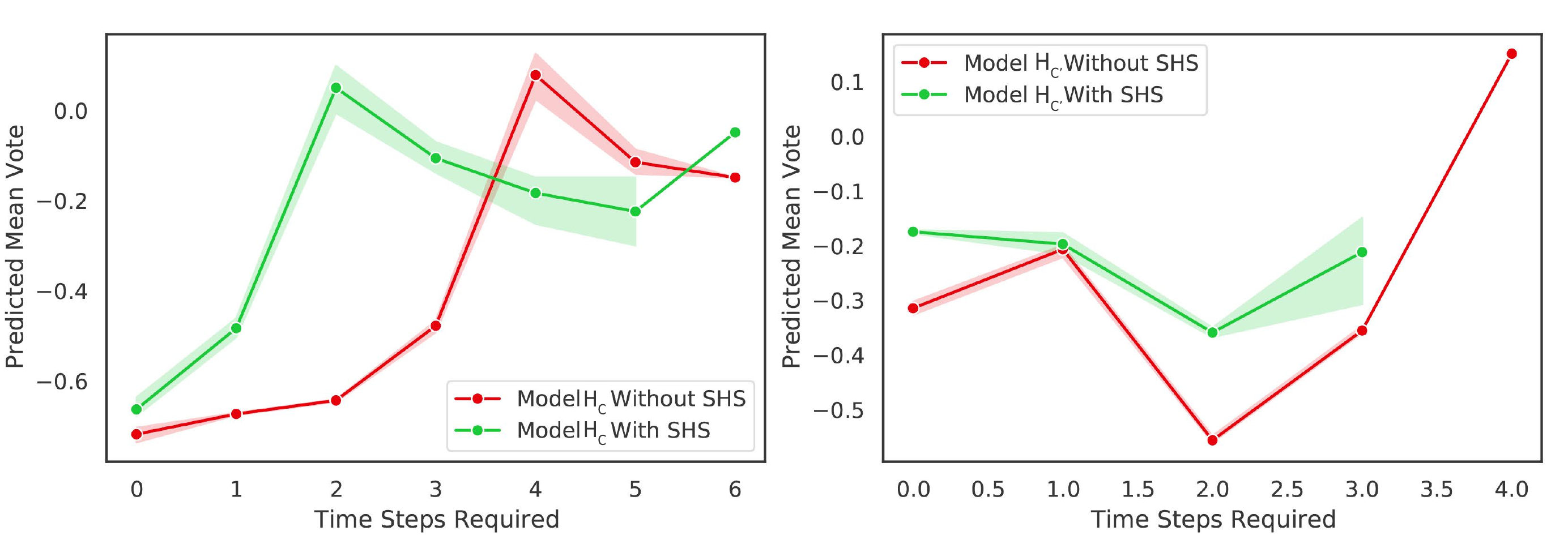}
    \caption{Experiment 6: Model $\mathcal{H_C}$ and Model $\mathcal{H_C}$' PMV trajectory while completing all activities {\color{green}with} and {\color{red}without} the SHS for PMV range [-0.25, 0.25]}
    \label{Exp_6_Timesteps}
\end{figure}
\begin{figure}[t]
    \centering
    \includegraphics[width=0.75\columnwidth]{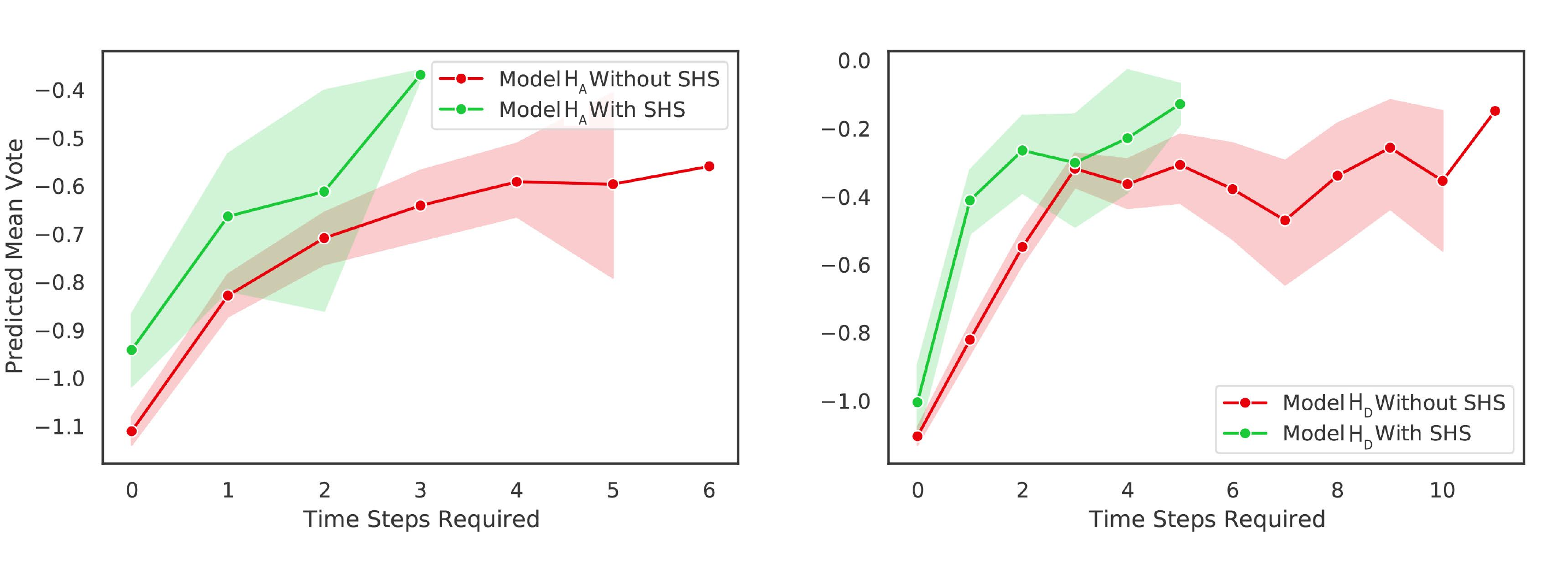}
    \caption{Experiment 7: Model $\mathcal{H_A}$ and Model $\mathcal{H_D}$ PMV trajectory while completing all activities {\color{green}with} and {\color{red}without} the SHS.}
    \label{Exp_7_Timesteps}
\end{figure}

\subsubsection{Experiment 4: $\mathcal{H}_A$ \& $\mathcal{H}_B$}
In this experiment, we consider a scenario of having two human models with similar thermal preferences pursuing their activities at the same time and an SHS trying to learn their preferences. Here we integrate Models $\mathcal{H_A}$ and $\mathcal{H_B}$ simultaneously into the SHS. Both models have a similar thermal preference for the same activities along with the same intrinsic reward function. Similar to Experiment 1, we keep the metabolic indices for Model $\mathcal{H}_A$ as $[1.0, 1.3, 1.8]$ for activity 0, activity 1, and activity 2, respectively, and $[1.10, 1.35, 1.75]$ for Model $\mathcal{H}_B$ for the same sequence of activities. The reward function for both models can be define by Eq.~\ref{H_A_Reward_Equationn}.

In Figures \ref{Exp4_All}(a) and (c), after convergence, both $\mathcal{H}_A$ and $\mathcal{H}_B$ are able to complete each activity without frequent switching (in blue). Both agents set the TH before starting their activities (Set\# in Figure \ref{Exp4_All}) similar to what we might expect a human would do in a real-life scenario. The activity-reward plot in Figures \ref{Exp4_All}(b) and (d) show that in the presence of the SHS, both models can still manage to pursue activities without any change in the activity sequence and with reduced time-steps required to change TH preference as shown in Figure \ref{Exp_4_Timesteps}. The models pursue the activity in which the metabolism index difference among the two is at a minimum, so that the SHS does not have to learn to anticipate their different TH preference. Without the SHS, Model $\mathcal{H}_A$ took a mean of 6 time-steps to set the TH and a mean of 3 time-steps with the SHS. Similarly, without SHS, Model $\mathcal{H}_B$ took a mean of 11 time-steps, while this was reduced to 5 time-steps with the SHS. Here, we can observe that Model $\mathcal{H}_B$ gives more feedback (11 times) while Model $\mathcal{H}_A$ gives less feedback (6 times). The SHS also receives the activity that the model is pursuing while changing the TH thus it becomes more biased towards Model $\mathcal{H}_B$ than Model $\mathcal{H}_A$ as it receives more feedback from $\mathcal{H}_B$ which results in a large difference between the time-steps with and without the SHS for $\mathcal{H}_B$ when compared with $\mathcal{H}_A$. With this experiment, we observe that the SHS can become biased towards the human model which provides more feedback than the other. In the end, we can conclude this experiment as follows: (1) Both Models $\mathcal{H}_A$ and $\mathcal{H}_B$ can pursue their activities together, (2) the SHS can learn each model's TH preference showing bias towards the model that gives more feedback, and (3) the performance of the two human models is generally not impacted when integrated with the SHS.

\subsubsection{Experiment 5: $\mathcal{H_A}$ \& $\mathcal{H_C}$} 
In this experiment, we look into the scenario of where a baseline model and model with different intrinsic reward function pursue activities simultaneously in the same environment. Here our aim is to learn whether the SHS can learn the comfort preference for each model without changing their behavior. To do so, we utilize Model $\mathcal{H_C}$ which was introduced in Experiment 3, along with Model $\mathcal{H_A}$. The reward functions for Model $\mathcal{H_A}$ and $\mathcal{H_C}$ is given by Eq.~\ref{H_A_Reward_Equationn} and Eq.~\ref{H_C_Reward_Equation} respectively.

The models pursue their activities with maximum comfort in the same environment while the SHS agent learns to anticipate their comfort policy. It is important to note that the metabolism for $\mathcal{H_A}$ and $\mathcal{H_C}$ have a moderate difference. For the Model $\mathcal{H_C}$, the reward function is given in Eq. \ref{H_C_Reward_Equation}, and its metabolism indices are $[1.80, 1.35, 1.15]$, while for Model $\mathcal{H_A}$, the reward function is given in Eq. \ref{H_A_Reward_Equationn}, and its metabolism indices are $[1.0, 1.3, 1.8]$. It is important to note that the comfortable PMV range for $\mathcal{H_A}$ is between -0.5 to 0.5, while  for $\mathcal{H_C}$ it is between -0.25 to 0.25, indicating higher thermal sensitivity (Eq.~\ref{Thermal_Score}).

In Figure \ref{Exp5_All}(a) (in blue), we see that the Model $\mathcal{H_A}$ learns to pursue its activities and set the optimal TH preference, with only a few task switching. Similarly, Model $\mathcal{H_C}$ in Figure \ref{Exp5_All}(c) also converges by completing its activities and spends a bit more time setting the TH at the beginning of the activity. With the SHS, both models converge without any switching, as shown in Figures \ref{Exp5_All}(b) and (d) (in blue). Referring back to Experiment 3, we observed that Model $\mathcal{H_C}$ showed behavior anomalies due to the different intrinsic reward function with respect to the baseline model, as well as the SHS failing to learn its comfort policy. Interestingly, here we observe that Model $\mathcal{H_C}$ pursues activities in parallel to Model $\mathcal{H_A}$ without any anomalies. A possible explanation is that in this experiment there is a second human model ($\mathcal{H_A}$), working well with the SHS, that may result in a more optimal behaviour. A deeper analysis reveals that having the human models share the same environment is what accounts for the largest difference between the two experiments (3 and 5). In the single human experiment, Models $\mathcal{H_A}$ and $\mathcal{H_C}$ need 21 and 26 time-steps respectively (on average) to adjust the TH to their preference. However, together (experiment 5), they only need 16 time-steps (7 and 9 respectively). When adding the SHS, the number of time-steps required to adjust the TH is down to 12 (6 for each model). Therefore, it seems that the absence of anomalies in behaviour of $\mathcal{H_C}$ is mostly due to the fact that it interacts well with $\mathcal{H_A}$, despite their different preferences.

By looking at Figures \ref{Exp5_All}(a) and (c), we can observe that the human models use a different activity order from each other which allows them to synchronize their activities with matching TH preferences. This can be seen in both cases, with and without the SHS. When comparing the metabolism rate of their activities at any time-step, we find that the difference between their indices is minimum. Accordingly, when $\mathcal{H_A}$ pursues \textit{Activity 0}, $\mathcal{H_C}$ pursues \textit{Activity 2}. Similarly when $\mathcal{H_A}$ pursues \textit{Activity 1}, $\mathcal{H_C}$ pursues \textit{Activity 1}. And finally when $\mathcal{H_A}$ pursues \textit{Activity 2}, $\mathcal{H_C}$ pursues \textit{Activity 0}. Interestingly, the same behavior can be observed when both models are integrated with the SHS. Having minimal difference in the metabolism rate results in similar TH preferences for their activities \cite{standard199255}. This in turn reduces the required time-steps by the models need to change the TH settings as when either of the models change the TH settings towards their preference, it also advances the other agent towards the state with optimal comfort. 

To further assess that it is the positive interaction between the two human models that eliminates the behavioural anomalies of Model $\mathcal{H_C}$, visible in Experiment 3, we rerun Experiment 5 with a more restrictive preference range ($d_{0.25}$ instead of $d_{0.50}$) for both models. The result is shown in Figure~\ref{fig:Experiment_5_Time Steps_0.25}. In this case, we can see that without the SHS, both models show more frequent task switching than in Figure~\ref{Exp5_All} with Model $\mathcal{H_C}$ being more affected. When the SHS is added, model $\mathcal{H_C}$ switches tasks even more frequently, as found in Experiment 3.


\begin{table}[t]
    \centering
    \caption{Experiment 4: MTS required to change TH, MR $\pm$ standard deviation for each model with and without SHS. }
      \begin{tabular}{|c|c|c|c|c|}
   \hline
    & \multicolumn{2}{c|}{Without SHS} & \multicolumn{2}{c|}{With SHS} \\
      & {MTS} & {MR} & {MTS} & {MR}\\
   \hline
    Model $\mathcal{H_A}$ & 6 & 290 $\pm$ 3.6 & 3 & 289 $\pm$  5.47\\  
    Model $\mathcal{H_B}$ & 11 & 290 $\pm$ 6.6 & 5 & 288 $\pm$  7.91\\
    \hline
    \end{tabular}
    
    \label{Table_Exp_4}
\end{table}

\begin{table}[!h]
    \centering
    \caption{Experiment 5: MTS required to change TH, MR $\pm$ standard deviation for each model with and without SHS. }
    \begin{tabular}{|c|c|c|c|c|}
   \hline
    & \multicolumn{2}{c|}{Without SHS} & \multicolumn{2}{c|}{With SHS} \\
      & {MTS} & {MR} & {MTS} & {MR}\\
   \hline
    Model $\mathcal{H_A}$ & 7 & 286 $\pm$ 2.36 & 6 & 276 $\pm$ 6.55\\  
    Model $\mathcal{H_C}$ & 9 & 287 $\pm$ 7.74 & 6 & 288 $\pm$ 5.26\\
    \hline
    \end{tabular}
    \label{Table_Exp_5}
\end{table}

\begin{table}[!h]
    \centering
    \caption{Experiment 6: MTS required to change TH, MR $\pm$ standard deviation for each model with and without SHS. Here the PMV range for both model is [-0.25, 0.25]}
       \begin{tabular}{|c|c|c|c|c|}
   \hline
    & \multicolumn{2}{c|}{Without SHS} & \multicolumn{2}{c|}{With SHS} \\
      & {MTS} & {MR} & {MTS} & {MR}\\
   \hline
    Model $\mathcal{H_C}$ & 6 & 287 $\pm$ 6.16 & 6 & 288 $\pm$  3.4\\  
    Model $\mathcal{H_C'}$ & 4 & 291 $\pm$ 6.30 & 3 & 292 $\pm$  3.3\\
    \hline
    \end{tabular}
    \label{Table_Exp_6}
\end{table}

\begin{table}[!h]
    \centering
    \caption{Experiment 7: MTS required to change TH, MR $\pm$ standard deviation for each model with and without SHS. }
    \begin{tabular}{|c|c|c|c|c|}
   \hline
    & \multicolumn{2}{c|}{Without SHS} & \multicolumn{2}{c|}{With SHS} \\
      & {MTS} & {MR} & {MTS} & {MR}\\
   \hline
    Model $\mathcal{H_A}$ & 6 & 283 $\pm$ 4.1 & 3 & 274 $\pm$  5.6\\  
    Model $\mathcal{H_D}$ & 11 & 284 $\pm$ 5.2 & 5 & 276 $\pm$  7.1\\
    \hline
    \end{tabular}
    \label{Table_Exp_7}
\end{table}

\subsubsection{Experiment 6: $\mathcal{H_C}$ \& $\mathcal{H_C'}$} In this experiment, we hypothesize a scenario where two simulated thermally sensitive models pursue activities within the same environment with the SHS at the same time. The intrinsic reward function for both models can be defined as Eq. \ref{H_C_Reward_Equation} which is different from the reward function of baseline models. Here we include Model $\mathcal{H_C}$ (from experiment 3) and Model $\mathcal{H_C'}$. Model $\mathcal{H_C'}$ is similar to $\mathcal{H_C}$, except that it has a slightly different thermal preference for each activity. For Model $\mathcal{H_C}$ the metabolism indices in $[1.80, 1.35, 1.15]$, while for Model Model $\mathcal{H_C'}$, it is $[1.75, 1.30, 1.15]$. We represent the thermal sensitivity by setting the PMV range between $-0.25$ and $0.25$ which is rewarded by $d_{0.25}$. In this range, the number of TH combinations for optimal comfort is 12, thus providing more options (as shown in Table \ref{Combination2}) where both human agents can feel comfortable. To obtain these combinations, we used the thermal comfort tool \cite{cbe} to get the TH combinations for the above mentioned PMV range, empirically the thermal step is set to 1 degree, and the humidity step is set to 5\%. Narrowing down the PMV range would also reduce the combinations of comfortable TH making it more difficult for the agents to have a common preference among the TH combinations. With this experiment, we aim to learn (1) if both thermally sensitive models can learn to pursue their activities and also set TH, and (2) if the SHS can learn the comfort policy for both models without any anomaly for the human agents.

In Figures \ref{Exp6_All}(a) and (b), we observe that Model $\mathcal{H_C}$ converges to an optimal policy to complete the activities without frequent switching. Then in Figures \ref{Exp6_All}(c) and (d), we observe that Model $\mathcal{H_C'}$ also converges to an optimal policy to complete the activities without frequent switching. The stability of both sensitive models is due to the high number of comfortable TH combinations and similar intrinsic reward function (Eq. ~\ref{H_C_Reward_Equation}). From Figure \ref{Exp_6_Timesteps}, we observe that in the PMV trajectory for $\mathcal{H_C}$ and $\mathcal{H_C'}$ the number of time-steps required to set the TH is reduced. Table \ref{Table_Exp_6} presents the mean time-steps (MTS), mean reward (MR), and its standard deviation, with and without the SHS. From Figure \ref{Exp6_All}(b) and (d) we observe that two that two thermally sensitive human models $\mathcal{H_C}$ and $\mathcal{H_C}$' can remain stable with the SHS without showing any behavior anomaly like frequent switching between activities and increase in time-steps to set TH.

\subsubsection{Experiment 7: $\mathcal{H_A}$ \& $\mathcal{H_D}$} In this experiment, we simulate a scenario where models have different intrinsic reward functions but similar thermal preferences. We use Model $\mathcal{H_D}$ from Experiment 3 whose intrinsic reward function can be expressed by Eq. \ref{H_C_Reward_Equation}. In Experiment 3, we observed behavior anomalies with Model $\mathcal{H_D}$ when integrated with the SHS as shown in Figure \ref{Exp_1_2_3_timestep} and Table \ref{Exp_123_MTS_MR}. Thus, with this simulated experiment, we aim to learn whether two human models can display behavioral anomalies where the models have different reward functions but similar thermal preference. To keep similar thermal preferences, we kept the metabolism indices for Model $\mathcal{H_A}$ and Model $\mathcal{H_D}$ as $[1.0, 1.3, 1.8]$ and $[1.15, 1.25, 1.85]$ respectively. 

Figures \ref{Exp7_All}(a) and (c) show the activity plots for $\mathcal{H_A}$ and $\mathcal{H_D}$ without the SHS (in blue). No occasional switching was observed because the comfortable TH options for $\mathcal{H_D}$ is a subset of the TH options for $\mathcal{H_A}$ as the thermal preference is similar resulting in pursuing same activities. From Table \ref{Table_Exp_7}, we observe that $\mathcal{H_D}$ gives more feedback (more time-steps) to the SHS, making the SHS biased towards the model. 

The activity performance for the two models is presented in Figures \ref{Exp7_All}(b) and (d) (in blue), where we observe that neither model encounters frequent switching, with or without the SHS. Even though the intrinsic reward function of the two models is different, the comfortable PMV range for Model $\mathcal{H_D}$ ($-0.25$ to $0.25$) lies within the PMV range of Model $\mathcal{H_A}$ ($-0.5$ to $0.5$), as well as having a similar thermal preference. Moreover, having a similar metabolism rate for the same activities results in similar thermal preference. Thus, attaining the comfortable TH state for Model $\mathcal{H_D}$ (the more sensitive model because of the constrictive PMV range) would also achieve a comfortable state for Model $\mathcal{H_A}$. With two human agents in the same environment, the number of time-steps for each model is reduced. This can be compared from Table \ref{Exp_123_MTS_MR} and Table \ref{Table_Exp_7} where for Model $\mathcal{H_D}$ as a single agent takes 26 time-steps with the SHS to set TH along with occasional switching, whereas with two-human agent, Model $\mathcal{H_D}$ only takes 5 time-steps. Therefore when human models with similar TH preference are integrated with the SHS, the SHS doesn't have to learn a different comfort policy for each model, thus the SHS's learned policy that provides comfort is similar for both human models. Although, bias towards the model that gives more feedback can also observed. Figure \ref{Exp_7_Timesteps} illustrates the improvements in $\mathcal{H_A}$ and $\mathcal{H_D}$ with the SHS, where the trajectory for $\mathcal{H_D}$ seems to have better improvement as compared to $\mathcal{H_A}$ due to $\mathcal{H_D}$ providing more feedback. The slope of the PMV curve improves with the SHS versus without it, achieving the comfortable TH within fewer time-steps. We can also observe from Figure \ref{Exp7_All} (in blue) that with the SHS, both models mostly switch to the same activities, thus having similar metabolisms most of the time to reduce the thermal difference and avoid time spent to change the TH.

\section{Conclusion}
We investigate the interaction between simulated human models and smart homes in the context of setting thermal preferences. To do so, we simulate human models based on hierarchical reinforcement learning capable of continuing and leaving a set of of predefined activities and also setting thermal preferences. Our smart home is based on Q-learning and learns the preferences of the human model for each activity. While the SHS could reduce the time spent by the human model toward setting the thermal preferences, we observe that human models with reward functions different than the one our smart home is trained with, show behavior anomalies such as frequent switching and an increase in the time required to set the thermal preferences. Similar results were found when two human models were sharing the same environment. Depending on the two selected human models, these models could see improved performance together and with the SHS. But again, a model different than the one for which the SHS was designed for could lead to frequent activity switching and frequent change to the temperature and humidity. Our study highlights the importance of validating and testing smart-home systems against wide variety of end-users to ensure that the pre-trained smart home systems and products perform as desired without negative impacts on the users.

\small
\bibliographystyle{abbrv}
\bibliography{reference}
\end{document}